\documentclass[journal]{IEEEtran}
\usepackage{cite}
\usepackage{amsmath,amssymb,amsfonts}
\usepackage{algorithmic}
\usepackage{graphicx}
\usepackage{algorithm,algorithmic}
\usepackage{hyperref}
\hypersetup{hidelinks}
\usepackage{textcomp}
\usepackage{balance} 

\newcommand{\R}{\mathbb R}
\newcommand{\X}{\mathcal X}
\newcommand{\T}{\mathcal T}

\newcommand{\G}{\mathcal G}
 
\newcommand{\dd}{\,\mathrm d}
\newcommand{\rec}{\mathrm{rec}}
\newcommand{\new}{\mathrm{new}}
\newcommand{\argmin}{\operatorname*{argmin}}
\newcommand{\matM}{\mathbf{M}}
\newcommand{\bfc}{\mathbf{c}}
\newcommand{\loc}{\mathrm{loc}}

\newtheorem{theorem}{Theorem}
\newtheorem{lemma}{Lemma}
\newtheorem{proposition}{Proposition}
\newtheorem{corollary}{Corollary}
\newtheorem{assumption}{Assumption}
\newtheorem{definition}{Definition}
\newtheorem{remark}{Remark}

\begin{document}
\title{Recoverability, Not Invariance: Certified Safety for Robots \\ at the Limits of Performance}
\author{Thomas A. Berrueta$^{1,2}$, Joshua D. Ibrahim$^{1}$, Gilchrist X. Johnson$^{1}$, John Lathrop$^{1}$, and Soon-Jo Chung$^{1}$
\thanks{$^{1}$These authors are with the Department of Computing and Mathematical Sciences, California Institute of Technology, Pasadena, CA 91125 USA (e-mail: \{jdibrahi, gxj, jlathrop, sjchung\}@caltech.edu).}
\thanks{$^{2}$This author is with the Department of Mechanical Engineering, Stanford University, Stanford, CA 94305 USA (e-mail: berrueta@stanford.edu).}
\thanks{This work has been submitted to the IEEE for possible publication. Copyright may be transferred without notice, after which this version may no longer be accessible.}
}

\maketitle

\begin{abstract}
Safety certification in robotics typically rests on forward invariance of the deployed controller, a property that demands high-fidelity models at the edge of the operating envelope, exactly where models are least reliable and robots perform best. As a result, many of the behaviors we most want robots to run, from learned policies to aggressive maneuvers at the limits of traction, cannot be certified at all. We introduce recoverability, a contraction-theoretic safety concept that resolves this tension by certifying a robot's ability to reengage a trusted recovery mode rather than the invariance of a deployed behavior. We prove that switching to the recovery mode from anywhere inside a certified tube guarantees exponential convergence, even under disturbances and model error. To decide when to switch, we derive the contraction insertion gradient, the first-order effect of a recovery insertion on the certificate, and embed it in a switching-time program in which feasibility alone guarantees safety and optimality only shapes performance. To supply the certificates, we develop sequential convex synthesis that monotonically grows the certified tube toward the exact viability kernel. We validate our approach in simulated quadrotor flight through uncertain obstacles, where it outperforms invariance-based safety filters, and in hardware deployments aboard a full-scale IndyNXT autonomous racecar, where it repeatedly recovers the car from an aggressive raceline at the limits of handling in the rain. Recoverability provides a certified exit strategy, extending formal safety guarantees to aggressive, learned, and experimental behaviors without requiring their invariance.

\textit{Supplementary Video---}\url{https://youtu.be/h1kTaiRAQEI}.
\end{abstract}

\begin{IEEEkeywords}
Robot Safety, Formal Methods in Robotics and Automation, Optimization and Optimal Control, Contraction Theory.
\end{IEEEkeywords}

\section{Introduction}
\label{sec:introduction}

\IEEEPARstart{R}{obots} are increasingly asked to operate near the limits of their capabilities. Legged robots run and recover from falls using policies learned in simulation~\cite{hwangbo2019learning}. Learned controllers race drones past human champions and to the physical limits of the platform~\cite{kaufmann2023champion,song2023reaching}. Automated vehicles exploit tire saturation to avoid collisions and to drive at the limits of adhesion~\cite{funke2017collision,kritayakirana2012autonomous,betz2022autonomous}. In each case, the most valuable behaviors lie near the boundary of the operating envelope, and improving performance means operating closer to that boundary or expanding the envelope itself, a practice known as envelope expansion. However, conventional safety certification penalizes operation at the boundary, treating peak performance as a hazard to be avoided.

\begin{figure*}[t]
    \centering
    \includegraphics[width=0.9\textwidth]{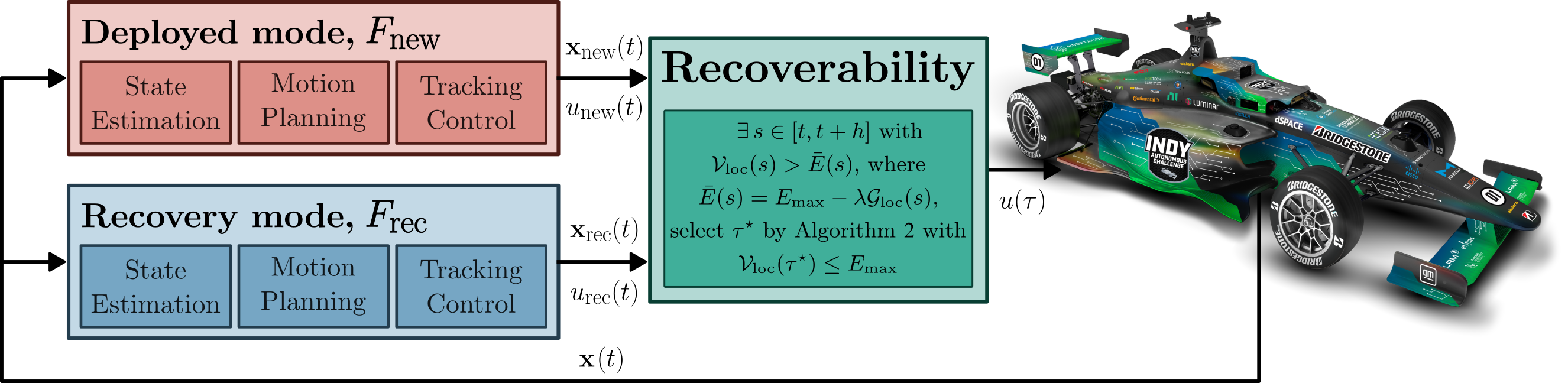}
    \caption{The recoverability architecture. Because recoverability is defined directly on closed-loop vector fields, each mode can be built from any part of an autonomy stack, including the state estimation, motion planning, and tracking control shown here, as well as policies or other learned components. The deployed mode $F_{\new}$ runs unmodified, and the recovery mode $F_{\rec}$ is the trusted behavior whose recovery tube is certified offline. The recoverability block summarizes Algorithm~\ref{alg:onboard}, which uses the predictions of both modes to schedule the handoff at time $\tau^\star$.}
    \label{fig:architecture}
\end{figure*}

Most formal safety frameworks are based on the principle of forward invariance~\cite{blanchini1999set}, in which a set of safe states is rendered invariant under the closed-loop dynamics. This principle underlies control barrier functions (CBFs)~\cite{ames2017control,ames2019control}, Hamilton--Jacobi (HJ) reachability~\cite{mitchell2005time,bansal2017hamilton}, viability theory~\cite{aubin1991viability}, and predictive safety filters~\cite{wabersich2021predictive,hsu2024safety}. However, establishing forward invariance of the deployed closed-loop system requires a dynamical model that can be trusted throughout the safe set. High-performance operation concentrates the robot in regimes such as tire saturation, contact, and degraded surfaces, where models are least reliable and data are scarce. As a result, these guarantees are weakest where they are needed most. Invariance-based frameworks also impose a direct cost on performance, since safety filters steer the system away from the boundary of the safe set even in minimally invasive formulations~\cite{fisac2019general,hsu2024safety}. Backup controller methods lighten the modeling burden by certifying a backup policy instead~\cite{gurriet2018towards,chen2021backup}, but they retain its filter and cost, and their membership certificates offer no convergence rate or notion of margin.

In this work, we show that robots do not need certified behavior to be certifiably safe. We formalize safety as \emph{recoverability}, the certified ability of a robot to fall back on a trusted recovery behavior. Recoverability moves the invariance requirement off the deployed controller and onto the recovery mode, so the deployed controller may be aggressive, learned, or experimental, with no contraction, invariance, or modeling requirement. Figure~\ref{fig:architecture} shows the resulting architecture, in which a recoverability monitor hands control from the deployed mode to the recovery mode at a certified time. Using contraction theory, we certify a tube around the recovery reference from which recovery converges exponentially. Monitoring safety then reduces to evaluating a closed-form Riemannian distance, and the guarantee degrades gracefully under perturbations. The contraction insertion gradient then tells the robot when to hand off, scheduling the switch through an optimization whose feasible solutions are all safe. As a result, robots can run behaviors that invariance-based safety cannot certify, requiring only that they are recoverable. 

The contributions of this article are as follows.
\begin{enumerate}
    \item We introduce recoverability, a contraction-theoretic safety concept that certifies a trusted recovery mode instead of the deployed controller, and prove that switching within a tube yields robust exponential convergence.
    \item We derive the contraction insertion gradient and use it to schedule switches through an optimization in which feasibility alone certifies recovery, and we synthesize certified tubes by sequential convex programming, benchmarking them against exact HJ viability.
    \item We demonstrate recoverability in quadrotor flight past uncertain obstacles, where it outperforms CBFs and backup CBFs, and onboard a full-scale IndyNXT autonomous racecar at the limits of handling in the rain.
\end{enumerate}
Our results and experiments are also shown and summarized in a supplementary video accompanying this manuscript.

The rest of the manuscript is structured as follows: Section~\ref{sec:related-works} positions recoverability against invariance-based and fallback-based safety. Section~\ref{sec:preliminaries} defines the closed-loop modes and recovery tubes. Section~\ref{sec:scheduling} presents our main results, proving recoverability from the tube, turning the certificate into an optimal switching decision, and evaluating it in quadrotor flight. Section~\ref{sec:certificate_synthesis} grows certified tubes through convex synthesis and benchmarks them against exact HJ viability. Section~\ref{sec:racecar} deploys the framework onboard a full-scale autonomous racecar, and Section~\ref{sec:discussion} concludes.

\section{Related Work}
\label{sec:related-works}

\begin{figure*}[t]
    \centering
    \includegraphics[width=0.9\textwidth]{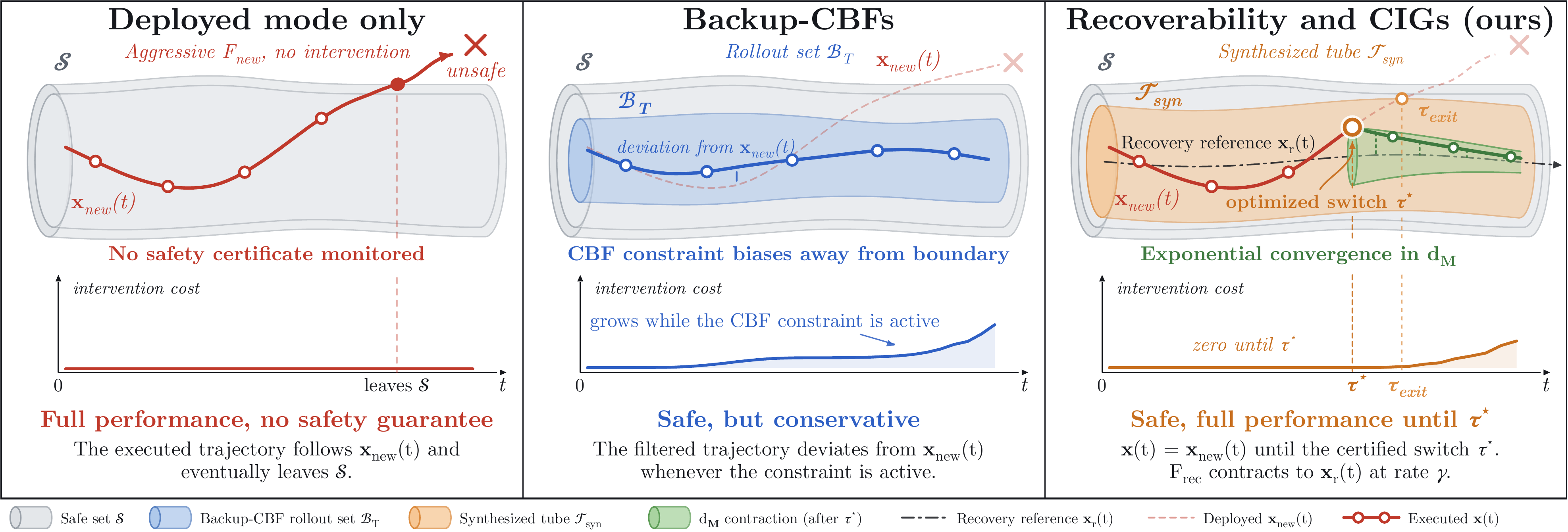}
    \caption{Safety and performance under three deployment strategies.
    In the left panel the aggressive deployed mode runs with no intervention, so the executed trajectory follows $\mathbf{x}_{\new}(t)$ at zero performance cost but eventually leaves the safe set $\mathcal S$ with no certificate to warn of the exit.
    In the center panel a backup-CBF filter keeps the state inside the rollout set $\mathcal B_T$, but the filtered trajectory deviates from $\mathbf{x}_{\new}(t)$ whenever the constraint is active, so the filter incurs a performance cost whenever the constraint is active.
    In the right panel our framework monitors the synthesized recovery tube $\T_{\rm syn}\subseteq \mathcal S$ and executes $\mathbf{x}_{\new}(t)$ untouched until the certified switch at $\tau^\star$, placed ahead of the latest certified time $\tau_{\rm exit}$, after which the recovery mode $F_{\rec}$ contracts to the reference $\mathbf{x}_r(t)$ exponentially in $d_{\matM}$, so the performance cost is zero until $\tau^\star$ and safety is guaranteed throughout.}
    \label{fig:comparison_panel}
\end{figure*}

The largest family of certification methods rests on forward invariance. HJ reachability computes the maximal safe set and answers the safety question exactly, but it requires a trusted global model and offline computation that scales exponentially with state dimension~\cite{mitchell2005time,bansal2017hamilton}. Control barrier functions reduce the online burden to a quadratic program, but synthesizing a valid barrier function is itself a difficult problem that scales poorly beyond modest dimensions~\cite{ames2017control,ames2019control}. Predictive safety filters replace offline synthesis with online optimization over a receding horizon, and they inherit the model requirements of the underlying predictions~\cite{wabersich2021predictive,hsu2024safety}. In every case the certified object is the deployed closed-loop system, so the modeling burden grows with the aggressiveness of the deployed behavior, and the filter itself biases operation away from the envelope boundary~\cite{fisac2019general}. Our framework instead inherits its guarantee from the recovery mode's contraction certificate for as long as a certified switch remains within reach, so the modeling burden falls on the recovery mode alone and the deployed behavior is never filtered at the boundary.

In our work, the certificate rests on contraction theory, which establishes incremental exponential convergence of trajectories under a Riemannian metric~\cite{lohmiller1998contraction,tsukamoto2021contraction}. Contraction-based tubes have been used for robust motion planning~\cite{singh2023robust}, and learned metrics extend these tools to data-driven settings~\cite{tsukamoto2021neural}, where contraction certifies the tracking performance of the deployed controller. Instead, we focus on certifying a recovery mode, and the minimizing geodesic of the metric supplies the gradient that drives the switching decision. Gradient formulas for switching times~\cite{xu2004optimal,egerstedt2006transition} and the mode insertion gradients behind sequential action control~\cite{ansari2016sequential} optimize performance objectives with the mode schedule as the design variable. The contraction insertion gradient is the insertion gradient of a safety certificate rather than a performance cost, so the resulting switching-time problem certifies safety through feasibility alone. Certificate synthesis has its own literature, from maximum-volume invariant ellipsoids~\cite{boyd1994linear} and sum-of-squares barrier certificates~\cite{prajna2004safety} to funnel libraries that precompute certified tubes for online composition~\cite{majumdar2017funnels}. Control contraction metrics make the search for a metric convex~\cite{manchester2017control}, but the geodesic distances their certificates require are rarely available in closed form. Our transport certificates combine the flexibility of these richer classes with the closed-form online test of an ellipsoid, and accepted sequential convex updates grow the tube monotonically while preserving certification.

Recoverability shares its premise with simplex and runtime-assurance architectures, which separate an unverified performance controller from a verified backup controller~\cite{seto1998simplex,hobbs2023runtime}. Recent work optimizes the supervisor and automates recoverability proofs~\cite{miller2024optimal,nigam2023recoverability}. The closest works to ours are backup controller methods~\cite{gurriet2018towards,gurriet2020scalable,chen2021backup} and their disturbance-robust extensions~\cite{vanwijk2024disturbance}, which fix a trusted fallback policy and certify safety online by forward integrating its dynamics and checking membership in an implicitly defined invariant set, the rollout set $\mathcal B_T$. This shrinks the modeling burden to the fallback policy alone, the same relocation our framework performs, but the certificate is a binary membership test with no convergence rate~\cite{chen2021backup}, and evaluating it requires integrating the fallback flow and its sensitivity matrix at every step. A quadratic program then blends the nominal and backup commands, so the executed trajectory tracks neither the intended behavior nor a certified fallback and pays a performance cost whenever the constraint is active. Our certificate instead guarantees convergence at an explicit rate with robustness to disturbances and model error, accommodates time-varying references, and lets the switch be scheduled ahead of time.

Figure~\ref{fig:comparison_panel} summarizes this landscape, with uncertified deployment on the left, a backup-CBF filter accumulating a performance penalty in the center, and our framework on the right, which executes the intended behavior untouched until a single certified switch. The net result is a framework that requires model fidelity only of the recovery mode, certifies safety online through a closed-form distance and a scalar switching-time problem, and constructs its certificates offline by sequential convex programming.

\section{Preliminaries}
\label{sec:preliminaries}

This section presents the objects on which the recoverability framework is built. We first specify the two closed-loop operational modes of the system, as well as the reference trajectory that defines recovery. We then develop the Riemannian machinery used to measure deviations between trajectories, and introduce the contraction metric and recovery tube through which recovery is certified. The metric is synthesized offline, the recovery energy built from its distance function is monitored online, and the gradient of that energy drives the switching decisions of Section~\ref{sec:scheduling}.

\subsection{Closed-Loop Modes and Reference Consistency}
\label{sec:modes}

We describe the deployed and recovery behaviors by their closed-loop vector fields $F_{\new},F_{\rec}:\X\times\R_{\ge0}\to\R^n$ on an open and connected state space $\X\subseteq\R^n$, each continuous in time and $C^1$ in state. Working at the level of vector fields keeps the framework agnostic to how $F_{\new}$ is obtained, whether from a receding-horizon planner, a learned policy, or manual tuning. Let $u_{\new}(\mathbf{x},t)$ and $u_{\rec}(\mathbf{x},t)$ be the corresponding control signals, $\mathcal U(\mathbf{x},t)$ the admissible control set, and $\mathcal S(t)=\{\mathbf{x}:h_i(\mathbf{x},t)\ge0,\ i=1,\ldots,N_h\}$ the safe set. All statements are restricted to intervals on which the relevant solutions exist uniquely and remain in $\X$.

Recovery is defined relative to a reference $\mathbf{x}_r:\R_{\ge0}\to\X$ that satisfies $\dot{\mathbf{x}}_r(t)=F_{\rec}(\mathbf{x}_r(t),t)$. This property, which we call reference consistency, makes the reference itself a trajectory of the trusted mode. The reference $\mathbf{x}_r(t)$ defines the recovery behavior and is not imposed as a tracking objective on the deployed mode. The deployed field $F_{\new}$ may arise from any admissible policy satisfying the stated regularity assumptions, including a controller tracking a separate performance reference or a policy acting directly on a task objective. Its execution need not converge toward $\mathbf{x}_r(t)$, since the pre-switch certification conditions concern continued recovery-tube membership and input admissibility, as specified in Section~\ref{sec:root_finding_switching}. 

For a switching time $\tau\ge0$, the single-switch execution is
\begin{equation}
    \dot{\mathbf{x}}(t) = \begin{cases}
        F_{\new}(\mathbf{x}(t),t), & t<\tau,\\
        F_{\rec}(\mathbf{x}(t),t), & t\ge\tau,
    \end{cases} \quad \mathbf{x}(\tau^+)=\mathbf{x}(\tau^-).
    \label{eq:single_switch_execution}
\end{equation}
Every switch we consider is of this form, and in Section~\ref{sec:scheduling} we develop a framework for optimally selecting $\tau$ to guarantee the system can safely recover as needed.

\subsection{Recovery Contraction Metric and Recovery Tube}
\label{sec:metric_tube}

Deviations from the reference are measured with a $C^1$ Riemannian metric $\matM:\X\times\R_{\ge0}\to\mathbb S_{++}^n$, under which a tangent vector $\delta\mathbf{x}\in T_{\mathbf{x}}\X$ has length $\|\delta\mathbf{x}\|_{\matM(\mathbf{x},t)}:=\sqrt{\delta\mathbf{x}^\top\matM(\mathbf{x},t)\delta\mathbf{x}}$. The Riemannian distance and the recovery energy are
\begin{align}
    d_{\matM}(\mathbf{x},\mathbf{y},t) &:=\inf_{\bfc\in\Gamma(\mathbf{x},\mathbf{y})} \int_0^1\left\|\frac{\partial\bfc}{\partial s}(s)\right\|_{\matM(\bfc(s),t)}\dd s, \label{eq:riemannian_distance} \\
    \mathcal V_{\matM}(\mathbf{x},t) &:=\frac12d_{\matM}^2(\mathbf{x},\mathbf{x}_r(t),t), \label{eq:monitored_recovery_energy}
\end{align}
where $\Gamma(\mathbf{x},\mathbf{y})$ consists of piecewise $C^1$ curves in $\X$ from $\mathbf{x}$ to $\mathbf{y}$. This distance measures the finite separation between trajectories using the same metric that measures infinitesimal displacements.

\begin{definition}[Recovery contraction metric]
\label{def:recovery_contraction_metric}
The metric $\matM$ is a recovery contraction metric for $F_{\rec}$ on $\T\subseteq\X\times\R_{\ge0}$ with rate $\gamma>0$ if, at every $(\mathbf{x},t)\in\T$, it satisfies $\underline m\mathbf I\preceq\matM(\mathbf{x},t)\preceq\overline m\mathbf I$ for constants $0<\underline m\le\overline m$ and
\begin{equation}
    \dot{\matM}_{\rec}+\mathbf A_{\rec}^\top\matM+\matM\mathbf A_{\rec} \preceq-2\gamma\matM,
    \label{eq:contraction_metric_condition}
\end{equation}
where $\mathbf A_{\rec}:=\partial F_{\rec}/\partial\mathbf{x}$ is the Jacobian of the recovery field and $\dot{\matM}_{\rec}:=\partial_t\matM+\sum_{i=1}^nF_{\rec,i}\,\partial\matM/\partial x_i$ is the rate of change of $\matM$ along $F_{\rec}$.
\end{definition}

Condition~\eqref{eq:contraction_metric_condition} requires infinitesimal deviations between recovery trajectories to decay at rate $\gamma$. The contraction region $\T$ is verified offline and need not cover the entire state space.

\begin{definition}[Contractive recovery tube]
\label{def:contractive_recovery_tube}
For $r>0$, the contractive recovery tube is
\begin{equation}
    \T_r:=\{(\mathbf{x},t)\in\X\times\R_{\ge0}: d_{\matM}(\mathbf{x},\mathbf{x}_r(t),t)\le r\}.
    \label{eq:recovery_tube}
\end{equation}
With $E_{\max}:=\tfrac12r^2$, the tube is equivalently the sublevel set $\mathcal V_{\matM}\le E_{\max}$ of the recovery energy.
\end{definition}

The tube collects the states from which recovery will be certified, and through our framework we can test membership online by comparing $\mathcal V_{\matM}$ against $E_{\max}$. Certification rests on the following two conditions, both verified offline for the trusted recovery mode. 

\begin{assumption}[Contracting recovery tube]
\label{ass:contracting_recovery_tube}
The metric $\matM$ is a recovery contraction metric for $F_{\rec}$ with rate $\gamma$ on an open set $\T\supset\T_r$, and $d_{\matM}(\mathbf{x},\mathbf{x}_r(t),t)$ is continuous in $(\mathbf{x},t)$ near $\T_r$. For every $(\mathbf{x},t)\in\T_r$, there exists a constant-speed minimizing geodesic $\bfc^\star(\cdot;\mathbf{x},t)$ from $\mathbf{x}_r(t)$ to $\mathbf{x}$.
\end{assumption}

\begin{assumption}[Safe recovery tube]
\label{ass:safe_recovery_tube}
For every $(\mathbf{x},t)\in\T_r$, the state and recovery input satisfy
\begin{equation*}
    \mathbf{x}\in\mathcal S(t), \qquad u_{\rec}(\mathbf{x},t)\in\mathcal U(\mathbf{x},t).
\end{equation*}
Every maximal solution of $\dot{\mathbf{x}}=F_{\rec}(\mathbf{x},t)$ that remains in $\T_r$ is forward complete, meaning it is defined for all future time.
\end{assumption}

\section{Certified Recovery via Mode Scheduling}
\label{sec:scheduling}

Given these preliminaries, in this section we present our main results. Theorems~\ref{thm:contractive_recoverability} and~\ref{thm:robust_recoverability} in Section~\ref{sec:recoverability} establish where switching is safe, proving that a switch from anywhere inside the tube guarantees exponential convergence to the reference, even under disturbances and model error. The remaining subsections determine when to switch, and do so optimally, by combining a closed-form gradient of the recoverability certificate, which we call the contraction insertion gradient, with a scalar switching-time program restricted to the first certified interval within the recovery tube. Theorem~\ref{thm:root_finding_certified_recoverability} in Section~\ref{sec:root_finding_switching} then shows that every feasible solution of this program certifies recovery, so feasibility alone guarantees safety. The section closes by testing the complete framework in quadrotor flight.

\subsection{Contractive Recoverability}
\label{sec:recoverability}

Every result in this section builds on a precise notion of when a robot can be recovered. A state is recoverable if a switch to the recovery mode is certified to return the robot safely to the reference from said state. This subsection shows that every state in the tube is recoverable, first for the nominal recovery mode and then under disturbances and reference error.

To show this, we track the distance $L(t):=d_{\matM}(\mathbf{x}(t),\mathbf{x}_r(t),t)$ from a trajectory $\mathbf{x}$ to the reference, whose rate of change at every instant is bounded by standard path-contraction conditions~\cite{lohmiller1998contraction,tsukamoto2021contraction}. Because the reference is itself a recovery trajectory, this is an incremental estimate between two solutions of $F_{\rec}$, specialized to the pair that defines the tube.

\begin{lemma}[Path contraction]
\label{lem:path_contraction}
Suppose Assumption~\ref{ass:contracting_recovery_tube} holds, and let $\mathbf{x}(t)$ solve $\dot{\mathbf{x}}=F_{\rec}(\mathbf{x},t)$. Whenever $(\mathbf{x}(t),t)\in\T_r$, the upper right Dini derivative of $L$ satisfies
\begin{equation}
    D^+L(t):=\limsup_{h\downarrow0}\frac{L(t+h)-L(t)}{h}\le-\gamma L(t).
    \label{eq:path_contraction}
\end{equation}
\end{lemma}

\begin{proof}
Every subsegment of a minimizing geodesic is itself minimizing, so the geodesic $\bfc^\star(\cdot;\mathbf{x}(t),t)$ of Assumption~\ref{ass:contracting_recovery_tube} satisfies $d_{\matM}(\bfc^\star(s;\mathbf{x}(t),t),\mathbf{x}_r(t),t)=sL(t)\le r$ and therefore lies in $\T_r\subset\T$. For small $h>0$, the recovery flow carries this geodesic into a path from $\mathbf{x}_r(t+h)$ to $\mathbf{x}(t+h)$ that remains in the open set $\T$. By~\eqref{eq:contraction_metric_condition}, each tangent vector of the path shrinks at rate $\gamma$ in the metric, so the path has length at most $e^{-\gamma h}L(t)$. The distance $L(t+h)$ is no greater than this length, which gives~\eqref{eq:path_contraction}.
\end{proof}

\begin{theorem}[Contractive recoverability]
\label{thm:contractive_recoverability}
Suppose Assumptions~\ref{ass:contracting_recovery_tube} and~\ref{ass:safe_recovery_tube} hold, and consider~\eqref{eq:single_switch_execution} with $(\mathbf{x}(\tau),\tau)\in\T_r$. Then, for all $t\ge\tau$,
\begin{equation}
    d_{\matM}(\mathbf{x}(t),\mathbf{x}_r(t),t) \le e^{-\gamma(t-\tau)}d_{\matM}(\mathbf{x}(\tau),\mathbf{x}_r(\tau),\tau)
    \label{eq:recoverability_distance}
\end{equation}
and
\begin{equation}
    \mathbf{x}(t)\in\mathcal S(t), \qquad u_{\rec}(\mathbf{x}(t),t)\in\mathcal U(\mathbf{x}(t),t).
    \label{eq:recoverability_safety}
\end{equation}
If the straight segment from $\mathbf{x}(\tau)$ to $\mathbf{x}_r(\tau)$ lies in a region where $\matM(\mathbf{x},\tau)\preceq\overline m\mathbf I$, then
\begin{equation*}
    \|\mathbf{x}(t)-\mathbf{x}_r(t)\| \le\sqrt{\frac{\overline m}{\underline m}}\, e^{-\gamma(t-\tau)}\|\mathbf{x}(\tau)-\mathbf{x}_r(\tau)\|.
\end{equation*}
\end{theorem}

\begin{proof}
After the switch, $\mathbf{x}$ follows $F_{\rec}$, so Lemma~\ref{lem:path_contraction} applies while the trajectory remains in $\T_r$. On the tube boundary it gives $D^+L\le-\gamma r<0$, which rules out an immediate exit, and the comparison lemma~\cite{khalil2002nonlinear} gives~\eqref{eq:recoverability_distance} while the trajectory remains in the tube. A later first exit would contradict $L(t)\le e^{-\gamma(t-\tau)}r<r$. Assumption~\ref{ass:safe_recovery_tube} then gives forward completeness and~\eqref{eq:recoverability_safety}. Finally, the lower metric bound along the minimizing geodesic gives $\sqrt{\underline m}\|\mathbf{x}(t)-\mathbf{x}_r(t)\|\le L(t)$, and the upper bound along the initial straight segment gives the Euclidean estimate.
\end{proof}

Theorem~\ref{thm:contractive_recoverability} places the invariance requirement on the recovery mode alone, so the guarantee after the switch requires no contraction or invariance of $F_{\new}$. Safety before the switch is addressed by the scheduling conditions of Section~\ref{sec:root_finding_switching}. The Euclidean estimate expresses the guarantee in the original coordinates, with the conditioning ratio $\overline m/\underline m$ setting its prefactor.

In practice, the recovery mode runs under disturbances and model error. We collect both in a perturbation $d$ of the recovery dynamics, $\dot{\mathbf{x}}=F_{\rec}(\mathbf{x},t)+d(\mathbf{x},t)$, where $d$ is continuous in time and locally Lipschitz in state, locally uniformly in time, and we bound it in the metric on the tube,
\begin{equation}
    \|d(\mathbf{x},t)\|_{\matM(\mathbf{x},t)}\le\bar d, \qquad (\mathbf{x},t)\in\T_r.
    \label{eq:disturbance_bound}
\end{equation}

\begin{theorem}[Robust contractive recoverability]
\label{thm:robust_recoverability}
Suppose Assumptions~\ref{ass:contracting_recovery_tube} and~\ref{ass:safe_recovery_tube} hold, with the forward completeness of Assumption~\ref{ass:safe_recovery_tube} imposed on the perturbed recovery dynamics, and let $d$ satisfy~\eqref{eq:disturbance_bound}. If $L(\tau)\le r$ and $\bar d\le\gamma r$, switching to the perturbed recovery dynamics keeps the trajectory in $\T_r$ and gives
\begin{equation*}
    L(t)\le e^{-\gamma(t-\tau)}L(\tau) +\frac{\bar d}{\gamma}\left(1-e^{-\gamma(t-\tau)}\right), \qquad t\ge\tau.
\end{equation*}
Thus $\limsup_{t\to\infty}L(t)\le\bar d/\gamma$, with state safety and admissible recovery inputs throughout.
\end{theorem}

\begin{proof}
While the trajectory lies in $\T_r$, the path in the proof of Lemma~\ref{lem:path_contraction} now ends at the state reached by the nominal recovery flow from $\mathbf{x}(t)$ rather than at $\mathbf{x}(t+h)$. A straight segment of length $h\|d(\mathbf{x}(t),t)\|_{\matM(\mathbf{x}(t),t)}+o(h)$ joins the two, so by~\eqref{eq:disturbance_bound} the resulting connecting path has length at most $e^{-\gamma h}L(t)+h\bar d+o(h)$, and
\begin{equation*}
    D^+L(t)\le-\gamma L(t)+\bar d.
\end{equation*}
At $L(t)=r$, the connecting path has length at most $r+h(\bar d-\gamma r)+o(h)\le r+o(h)$, because $\bar d\le\gamma r$. By the lower metric bound, the Euclidean distance from $\mathbf{x}(t+h)$ to the next tube slice, divided by $h$, therefore tends to zero. This is Nagumo's tangency condition~\cite{blanchini1999set}. Continuity of the distance makes the tube locally closed, and uniqueness of solutions gives forward invariance, including when $\bar d=\gamma r$. Comparison then yields the stated bound for all $t\ge\tau$, and Assumption~\ref{ass:safe_recovery_tube} gives the remaining conclusions.
\end{proof}

On the tube boundary, recovery shrinks the distance at rate $\gamma r$, while the disturbance can grow it at rate at most $\bar d$. The quantity $\gamma r$ is therefore the robustness margin of the tube, and the condition $\bar d\le\gamma r$ requires the disturbance to fit within it.

\begin{corollary}[Robustness to an infeasible reference]
\label{cor:infeasible_reference}
Under the hypotheses of Theorem~\ref{thm:robust_recoverability}, replace reference consistency by the residual $\mathbf r(t):=F_{\rec}(\mathbf{x}_r(t),t)-\dot{\mathbf{x}}_r(t)$ of a $C^1$ reference. While the trajectory remains in $\T_r$,
\begin{equation*}
    D^+L(t)\le-\gamma L(t)+\bar d +\|\mathbf r(t)\|_{\matM(\mathbf{x}_r(t),t)}.
\end{equation*}
If $\bar r:=\sup_{t\ge\tau}\|\mathbf r(t)\|_{\matM(\mathbf{x}_r(t),t)}<\infty$, comparison gives, on the same interval,
\begin{equation}
    L(t)\le e^{-\gamma(t-\tau)}L(\tau) +\frac{\bar d+\bar r}{\gamma}\left(1-e^{-\gamma(t-\tau)}\right).
    \label{eq:infeasible_reference_bound}
\end{equation}
If $L(\tau)\le r$ and $\bar d+\bar r\le\gamma r$, the trajectory remains in $\T_r$,~\eqref{eq:infeasible_reference_bound} holds for all $t\ge\tau$, and $\limsup_{t\to\infty}L(t)\le(\bar d+\bar r)/\gamma$, with state safety and admissible recovery inputs throughout.
\end{corollary}

\begin{proof}
In the connecting path of the proof of Theorem~\ref{thm:robust_recoverability}, the evolved geodesic now starts at the nominally evolved reference rather than at $\mathbf{x}_r(t+h)$. Joining the two by a second straight segment adds a length whose ratio to $h$ tends to $\|\mathbf r(t)\|_{\matM(\mathbf{x}_r(t),t)}$, which gives the stated inequality. The comparison and tangency arguments of Theorem~\ref{thm:robust_recoverability} then apply with $\bar d$ replaced by $\bar d+\bar r$.
\end{proof}

Corollary~\ref{cor:infeasible_reference} accounts for approximately consistent references such as the precomputed raceline in Section~\ref{sec:racecar}. The estimates hold on finite intervals where the hypotheses are verified and solutions exist. Guarantees for all future time additionally require continued validity of the hypotheses and forward completeness.

\subsection{Recovery Energy and Contraction Insertion Gradient}
\label{sec:cig}

Theorems~\ref{thm:contractive_recoverability} and~\ref{thm:robust_recoverability} establish where switching is safe, and checking it online requires only a single evaluation of the recovery energy~\eqref{eq:monitored_recovery_energy}, rather than computation of a viability kernel or solution of an HJ equation. Deciding when to switch turns this monitoring into optimization, which also requires the certificate to be differentiable.

\begin{assumption}[Differentiable recovery energy]
\label{ass:smooth_recovery_energy}
On the operating set $\mathcal D\subseteq \X\times\R_{\ge0}$, each pair $(\mathbf{x}_r(t),\mathbf{x})$ is connected by a unique constant-speed minimizing geodesic $\bfc^\star(\cdot;\mathbf{x},t)$ from $\mathbf{x}_r(t)$ to $\mathbf{x}$ for the metric $\matM(\cdot,t)$. Moreover, $\bfc^\star$ and $\partial\bfc^\star/\partial s$ depend $C^1$-smoothly on $(\mathbf{x},t)$. Thus, $\mathcal V_{\matM}\in C^1(\mathcal D)$.
\end{assumption}

This is a local normal-neighborhood condition for the time-dependent metric $\matM(\cdot,t)$, and it excludes cut-locus points where the squared distance is not differentiable. For $(\mathbf{x},t)\in\mathcal D$, let $\rho_{\matM}(\mathbf{x},t):=\nabla_{\mathbf{x}}\mathcal V_{\matM}(\mathbf{x},t)$ denote the gradient of the recovery energy.

\begin{lemma}[Endpoint gradient of recovery energy]
\label{lem:endpoint_gradient}
Under Assumption~\ref{ass:smooth_recovery_energy},
\begin{equation}
    \rho_{\matM}(\mathbf{x},t)
    =
    \matM(\mathbf{x},t)\frac{\partial \bfc^\star}{\partial s}(1;\mathbf{x},t).
    \label{eq:endpoint_gradient_formula}
\end{equation}
\end{lemma}

\begin{proof}
Fix $t$. For notational compactness write $q(s):=\bfc^\star(s;\mathbf{x},t)$ and $q_s(s):=\partial\bfc^\star(s;\mathbf{x},t)/\partial s$. Since $q$ is the minimizing geodesic from $\mathbf{x}_r(t)$ to $\mathbf{x}$,
\begin{equation}
    \mathcal V_{\matM}(\mathbf{x},t)
    =
    \frac12\int_0^1
    q_s(s)^\top \matM(q(s),t)q_s(s)\,\dd s.
    \label{eq:energy_along_min_geodesic}
\end{equation}
Let $\mathbf{x}_\epsilon=\mathbf{x}+\epsilon \mathbf{v}$, and let $q_\epsilon$ be the $C^1$ family of minimizing geodesics from $\mathbf{x}_r(t)$ to $\mathbf{x}_\epsilon$. Define $\eta(s):=\left.\partial q_\epsilon(s)/\partial\epsilon\right|_{\epsilon=0}$, so that $\eta(0)=0$ and $\eta(1)=\mathbf{v}$. Differentiating~\eqref{eq:energy_along_min_geodesic} gives
\begin{align}
    \left.\frac{\dd}{\dd\epsilon}\right|_{\epsilon=0}
    \mathcal V_{\matM}( &\mathbf{x}_\epsilon,t)
    =
    \int_0^1
    \eta_s^\top \matM(q,t)q_s\,\dd s
    \nonumber\\
    &\quad+
    \frac12\int_0^1
    q_s^\top
    \left(
    \sum_{i=1}^n
    \frac{\partial \matM}{\partial x_i}(q,t)\eta_i
    \right)
    q_s\,\dd s .
    \label{eq:first_variation_expanded}
\end{align}
Integrating the first integral by parts yields
\begin{equation}
    \int_0^1\eta_s^\top \matM q_s\,\dd s
    =
    \left[\eta^\top \matM q_s\right]_{0}^{1}
    -
    \int_0^1
    \eta^\top\frac{\dd}{\dd s}(\matM q_s)\,\dd s.
    \label{eq:first_variation_parts}
\end{equation}
Combining~\eqref{eq:first_variation_expanded} and~\eqref{eq:first_variation_parts},
\begin{align*}
    \left.\frac{\dd}{\dd\epsilon}\right|_{\epsilon=0}
    &\mathcal{V}_{\matM}(\mathbf{x}_\epsilon,t)
    =
    \mathbf{v}^\top \matM(\mathbf{x},t)q_s(1)
    \\
    &+
    \int_0^1
    \eta^\top
    \left[
    -\frac{\dd}{\dd s}(\matM q_s)
    +
    \frac12
    \nabla_q\!\left(q_s^\top \matM(q,t)q_s\right)
    \right]\dd s.
\end{align*}
The bracket is the Euler--Lagrange expression of the integral in~\eqref{eq:energy_along_min_geodesic}, which vanishes along the geodesic $q$, so
\begin{equation*}
    \left.\frac{\dd}{\dd\epsilon}\right|_{\epsilon=0}
    \mathcal V_{\matM}(\mathbf{x}_\epsilon,t)
    =
    \mathbf{v}^\top \matM(\mathbf{x},t)q_s(1).
\end{equation*}
Since this holds for every endpoint perturbation $\mathbf{v}$, the coordinate gradient is $\matM(\mathbf{x},t)q_s(1)$, which proves~\eqref{eq:endpoint_gradient_formula}.
\end{proof}

Consequently, Lemma~\ref{lem:endpoint_gradient} gives the closed-form gradient of the recoverability certificate $\mathcal V_{\matM}$ on the normal-neighborhood domain of Assumption~\ref{ass:smooth_recovery_energy}. The gradient is a byproduct of the geodesic computation already performed to evaluate the certificate, so no adjoint or sensitivity integration is needed to obtain it. Lastly, to define the contraction insertion gradient, let the mode discrepancy be $\Delta F(\mathbf{x},t):=F_{\rec}(\mathbf{x},t)-F_{\new}(\mathbf{x},t)$.

\begin{definition}[Contraction insertion gradient (CIG)]
\label{def:cig}
The contraction insertion gradient is
\begin{equation}
    \G_{\matM}(\mathbf{x},t)
    :=
    \rho_{\matM}(\mathbf{x},t)^\top\Delta F(\mathbf{x},t).
    \label{eq:cig_definition}
\end{equation}
\end{definition}

Equivalently, $\G_{\matM}=D_{\mathbf{x}}\mathcal V_{\matM}[\Delta F]$ is the directional derivative of the certified recoverability value $\mathcal V_{\matM}$ along the vector field difference $F_{\rec}-F_{\new}$. When $\G_{\matM}<0$, an infinitesimal insertion of the recovery mode decreases the recovery energy relative to continuing with the deployed mode, and when $\G_{\matM}>0$ the insertion is momentarily counterproductive. The CIG is therefore the mode insertion gradient from optimal control of switched systems~\cite{egerstedt2006transition,ansari2016sequential} applied to a safety certificate rather than a performance cost.

\begin{corollary}[Endpoint covector for the insertion derivative]
\label{cor:endpoint_covector}
Under Assumption~\ref{ass:smooth_recovery_energy},
\begin{equation*}
    \G_{\matM}(\mathbf{x},t)
    =
    \left[
    \matM(\mathbf{x},t)
    \frac{\partial \bfc^\star}{\partial s}(1;\mathbf{x},t)
    \right]^\top
    \Delta F(\mathbf{x},t).
\end{equation*}
As such, the covector multiplying the vector field variation in the first-order insertion derivative is the endpoint covector of the minimizing geodesic for $\mathcal V_{\matM}$.
\end{corollary} 

\begin{proof}
Lemma~\ref{lem:endpoint_gradient} gives $\rho_{\matM}=\matM\,\partial_s\bfc^\star(1;\mathbf{x},t)$. Substituting this identity into Definition~\ref{def:cig} proves the formula.
\end{proof} 

The CIG is a first-order object, and the next result verifies that it correctly captures the effect of finite insertions. To describe such insertions, let $\phi_{\alpha}(t_1,t_0,\mathbf{x}_0)$ denote the flow map of mode $\alpha\in\{\new,\rec\}$, namely the value at time $t_1$ of the solution of $\dot{\mathbf{x}}=F_{\alpha}(\mathbf{x},t)$ initialized at $\mathbf{x}(t_0)=\mathbf{x}_0$. Inserting the recovery mode at a candidate time $\tau$ for a duration $\lambda>0$ carries the state $\mathbf{x}(\tau)$ to $\phi_{\rec}(\tau+\lambda,\tau,\mathbf{x}(\tau))$, while remaining in the deployed mode carries it to $\phi_{\new}(\tau+\lambda,\tau,\mathbf{x}(\tau))$. The interval $[\tau,\tau+\lambda]$ is used to compare the two predicted endpoints and score the switching time $\tau$, as opposed to a recovery dwell time.

\begin{proposition}[Finite insertion identity]
\label{prop:finite_insertion}
Assume $\mathcal V_{\matM}\in C^1(\mathcal D)$, with $\mathcal V_{\matM}$ as defined in \eqref{eq:monitored_recovery_energy}. Let
\begin{align}
    \mathbf z_{\rec}(\lambda)
    :=
    \phi_{\rec}(\tau+\lambda,\tau,\mathbf{x}(\tau)),
    \nonumber\\
    \mathbf z_{\new}(\lambda)
    :=
    \phi_{\new}(\tau+\lambda,\tau,\mathbf{x}(\tau)).
    \label{eq:inserted_endpoints}
\end{align}
Let $\chi_\lambda:[0,1]\to\X$ be any $C^1$ curve satisfying
\begin{equation*}
    \chi_\lambda(0)=\mathbf z_{\new}(\lambda),
    \qquad
    \chi_\lambda(1)=\mathbf z_{\rec}(\lambda),
\end{equation*}
for $(\chi_\lambda(a),\tau+\lambda)\in\mathcal D$. Then the finite insertion energy difference is
\begin{align}
    &\mathcal V_{\matM}(\mathbf z_{\rec}(\lambda),\tau+\lambda)
    -
    \mathcal V_{\matM}(\mathbf z_{\new}(\lambda),\tau+\lambda)
    \nonumber\\
    &\quad=
    \int_0^1
    \rho_{\matM}(\chi_\lambda(a),\tau+\lambda)^\top \cdot
    \frac{\partial\chi_\lambda}{\partial a}(a)
    \dd a.
    \label{eq:finite_insertion_identity}
\end{align}
\end{proposition}

\begin{proof}
For fixed $\lambda$, define $\nu_\lambda(a):=\mathcal V_{\matM}(\chi_\lambda(a),\tau+\lambda)$. The endpoint conditions and the fundamental theorem of calculus give
\begin{align*}
    &\mathcal V_{\matM}(\mathbf z_{\rec}(\lambda),\tau+\lambda)
    -
    \mathcal V_{\matM}(\mathbf z_{\new}(\lambda),\tau+\lambda)
    \\
    &\qquad=
    \int_0^1
    \frac{\dd}{\dd a}
    \mathcal V_{\matM}(\chi_\lambda(a),\tau+\lambda)
    \dd a.
\end{align*}
Since the time is fixed with respect to $a$, chain rule gives
\begin{equation*}
    \frac{\dd}{\dd a}
    \mathcal V_{\matM}(\chi_\lambda(a),\tau+\lambda)
    =
    \rho_{\matM}(\chi_\lambda(a),\tau+\lambda)^\top
    \frac{\partial\chi_\lambda}{\partial a}(a),
\end{equation*}
which proves \eqref{eq:finite_insertion_identity}.
\end{proof}

Proposition~\ref{prop:finite_insertion} expresses the effect of a finite insertion as a line integral of the certificate gradient between the two predicted endpoints. Letting the insertion duration shrink to zero collapses this integral onto a single point, and what remains is the CIG.

\begin{corollary}[Infinitesimal insertion derivative]
\label{cor:infinitesimal_insertion_derivative}
Assume the hypotheses of Proposition~\ref{prop:finite_insertion} and that the flow maps defining \eqref{eq:inserted_endpoints} are differentiable in $\lambda$ at $\lambda=0$.
Then
\begin{equation*}
\begin{aligned}
    \left.
    \frac{\dd}{\dd\lambda}
    \right|_{\lambda=0^+}
    [
    \mathcal V_{\matM}(\mathbf z_{\rec}&(\lambda),\tau+\lambda)
    -
    \mathcal V_{\matM}(\mathbf z_{\new}(\lambda),\tau+\lambda)
    ] \\
    &=
    \G_{\matM}(\mathbf{x}(\tau),\tau).
\end{aligned}
\end{equation*}
\end{corollary}

\begin{proof}
At $\lambda=0$, $\mathbf{z}_{\rec}=\mathbf{z}_{\new}=\mathbf{x}(\tau)$, while $\dot{\mathbf{z}}_{\rec}(0^+)=F_{\rec}(\mathbf{x}(\tau),\tau)$ and $\dot{\mathbf{z}}_{\new}(0^+)=F_{\new}(\mathbf{x}(\tau),\tau)$. The two explicit $\partial_t\mathcal V_{\matM}$ terms cancel, leaving $\rho_{\matM}^\top(F_{\rec}-F_{\new})$ at $(\mathbf{x}(\tau),\tau)$, which is \eqref{eq:cig_definition}.
\end{proof}

Consequently, for any $\tau$ satisfying Corollary~\ref{cor:infinitesimal_insertion_derivative},
\begin{align}
    &\mathcal V_{\matM}(\mathbf z_{\rec}(\lambda),\tau+\lambda)
    -
    \mathcal V_{\matM}(\mathbf z_{\new}(\lambda),\tau+\lambda)
    \nonumber\\
    &\qquad=
    \lambda\G_{\matM}(\mathbf{x}(\tau),\tau)+o(\lambda)
    \qquad\text{as }\lambda\downarrow0.
    \label{eq:finite_insertion_first_order}
\end{align} 

In summary, this subsection establishes the CIG as the exact first-order model of a recovery insertion. Lemma~\ref{lem:endpoint_gradient} supplies its gradient in closed form, and Proposition~\ref{prop:finite_insertion} and Corollary~\ref{cor:infinitesimal_insertion_derivative} confirm that it governs finite insertions to first order. The remainder of this section certifies continued operation of the deployed mode and then selects the optimal switching time.

\subsection{Tube-Preserving Envelope Expansion}

The CIG compares energy rates at a single instant. This subsection uses second-order information to certify persistence of the deployed mode over a finite lookahead, which is the sense in which the framework supports aggressive operation. For a mode $\alpha\in\{\new,\rec\}$, define the energy rate along $\dot{\mathbf{x}}=F_\alpha(\mathbf{x},t)$ by
\begin{equation}
    \dot{\mathcal V}_{\matM,\alpha}(\mathbf{x},t)
    :=
    \frac{\partial \mathcal V_{\matM}}{\partial t}(\mathbf{x},t)
    +
    \rho_{\matM}(\mathbf{x},t)^\top F_\alpha(\mathbf{x},t).
    \label{eq:energy_rate}
\end{equation} 
The partial derivative includes the explicit time dependence of the metric and the reference evolution $\mathbf{x}_r(t)$. Importantly, the CIG in Definition~\ref{def:cig} is the difference between the instantaneous energy rates of the two modes.
\begin{equation*}
    \dot{\mathcal V}_{\matM,\rec}
    -
    \dot{\mathcal V}_{\matM,\new}
    =
    \G_{\matM}.
\end{equation*}

Then, assuming $\dot{\mathcal V}_{\matM,\new}$ is $C^1$ on $\mathcal D$, the second derivative of the recovery energy along the new mode is
\begin{equation*}
\begin{aligned}
    \ddot{\mathcal V}_{\matM,\new}(\mathbf{x},t)
    &={}
    \frac{\partial\dot{\mathcal V}_{\matM,\new}}{\partial t}(\mathbf{x},t)
    \\
    &\quad+
    \nabla_{\mathbf{x}}\dot{\mathcal V}_{\matM,\new}(\mathbf{x},t)^\top
    F_{\new}(\mathbf{x},t).
\end{aligned}
\end{equation*}
For a lookahead $h>0$, suppose there exists $L_{\mathcal V}>0$ such that every trajectory segment under $F_{\new}$ contained in $\mathcal D$ satisfies
\begin{equation}
    \left|
    \ddot{\mathcal V}_{\matM,\new}(\mathbf{x}(s),s)
    \right|
    \le
    L_{\mathcal V},
    \qquad s\in[t,t+h].
    \label{eq:second_order_energy_bound}
\end{equation} 
If $\ddot{\mathcal V}_{\matM,\new}$ is continuous on a compact set containing all deployed segments under consideration, any certified upper bound on $\left|\ddot{\mathcal V}_{\matM,\new}\right|$ over that set is a valid choice of $L_{\mathcal V}$. 
\begin{definition}[One-step recoverability margin]
\label{def:one_step_margin}
For a lookahead duration $h>0$, define
\begin{align}
    m^{\matM}_h(\mathbf{x},t)
    :=
     \ &E_{\max}
    -
    \mathcal V_{\matM}(\mathbf{x},t) \notag \\
    &-
    h\dot{\mathcal V}_{\matM,\new}(\mathbf{x},t)
    -
    \frac12L_{\mathcal V}h^2.
    \label{eq:one_step_margin}
\end{align}
\end{definition}

\begin{definition}[Tube-preserving admissible set]
\label{def:tube_preserving_admissible_set}
The tube-preserving admissible set over lookahead $h$ is
\begin{equation} 
    \Omega^{\matM}_h
    :=
    \left\{
    (\mathbf{x},t):
    \mathcal V_{\matM}(\mathbf{x},t)\le E_{\max},
    \quad
    m^{\matM}_h(\mathbf{x},t)\ge0
    \right\}.
    \label{eq:omega_h}
\end{equation}
\end{definition}

\begin{proposition}[One-step tube preservation]
\label{prop:tube_preservation}
If $(\mathbf{x}(t),t)\in\Omega^{\matM}_h$ as defined by \eqref{eq:omega_h} and the trajectory under $F_{\new}$ remains in $\mathcal D$ on $[t,t+h]$, then
\begin{equation}
    \mathcal V_{\matM}(\mathbf{x}(t+\theta),t+\theta)
    \le
    E_{\max},
    \qquad
    \theta\in[0,h].
\end{equation}
\end{proposition}

\begin{proof} 
For any $\theta\in[0,h]$, Taylor's theorem and \eqref{eq:second_order_energy_bound} give
\begin{align*}
    \mathcal V_{\matM}(\mathbf{x}(t+\theta),t+\theta)
    \le \mathcal V_{\matM}(&\mathbf{x}(t),t)
    \\ 
    &+
    \theta\dot{\mathcal V}_{\matM,\new}(\mathbf{x}(t),t)
    +
    \frac12L_{\mathcal V}\theta^2.
\end{align*}
The scalar function
\begin{equation*}
    q(\theta)
    :=
    \mathcal V_{\matM}(\mathbf{x}(t),t)
    +
    \theta\dot{\mathcal V}_{\matM,\new}(\mathbf{x}(t),t)
    +
    \frac12L_{\mathcal V}\theta^2
\end{equation*}
is convex in $\theta$. For $0\le\theta\le h$, convexity gives
\begin{equation*}
    q(\theta)
    \le
    \left(1-\frac{\theta}{h}\right)q(0)
    +
    \frac{\theta}{h}q(h).
\end{equation*}
Definition~\ref{def:tube_preserving_admissible_set}, through \eqref{eq:one_step_margin} and \eqref{eq:omega_h}, gives $q(0)\le E_{\max}$ and $q(h)\le E_{\max}$. Hence $q(\theta)\le E_{\max}$ for all $\theta\in[0,h]$.
\end{proof} 

\begin{corollary}[Largest certified lookahead] 
\label{cor:largest_certified_continuation}
Suppose $\mathcal V_{\matM}(\mathbf{x},t)\le E_{\max}$ and the same bound $L_{\mathcal V}>0$ in \eqref{eq:second_order_energy_bound} is valid over every deployed trajectory segment under consideration. The largest $h\ge0$ for which the margin \eqref{eq:one_step_margin} is nonnegative is
\begin{equation}
\begin{aligned}
    \frac{1}{L_{\mathcal V}}
    \bigg[&-\dot{\mathcal V}_{\matM,\new}(\mathbf{x},t)\\
    &+\sqrt{
    \dot{\mathcal V}_{\matM,\new}(\mathbf{x},t)^2
    +2L_{\mathcal V}
    \bigl(E_{\max}-\mathcal V_{\matM}(\mathbf{x},t)\bigr)
    }\bigg].
\end{aligned}
    \label{eq:largest_certified_continuation}
\end{equation}
Every shorter lookahead is certified by Proposition~\ref{prop:tube_preservation}, provided its deployed trajectory remains in $\mathcal D$.
\end{corollary} 

\begin{proof}
The condition $m^{\matM}_h(\mathbf{x},t)\ge0$ is a scalar quadratic inequality in $h$. Its largest nonnegative root is \eqref{eq:largest_certified_continuation}. The inequality holds between zero and that root because $m^{\matM}_0=E_{\max}-\mathcal V_{\matM}(\mathbf{x},t)\ge0$ and the coefficient of $h^2$ is negative.
\end{proof}

Proposition~\ref{prop:tube_preservation} and Corollary~\ref{cor:largest_certified_continuation} present a mechanism for envelope expansion. As long as the margin $m^{\matM}_h$ remains nonnegative, the deployed mode may continue without losing the recovery certificate, so aggression is limited by a measurable budget rather than by a fixed conservative envelope. Corollary~\ref{cor:largest_certified_continuation} gives the largest certified lookahead directly from the measured energy, its rate, and the verified curvature bound.

\subsection{Certified Recovery via Switching-Time Optimization}
\label{sec:root_finding_switching}

To conclude with the development of our recoverability framework, the switching time itself must be selected. We combine a finite mode insertion objective, which scores current recoverability and the improvement caused by recovery relative to continuing with the deployed mode, with the tube feasibility condition used in Theorem~\ref{thm:root_finding_certified_recoverability}. Fix an insertion duration $\lambda>0$, and let $\mathbf{x}_{\new}(\cdot)$ denote the solution of $\dot{\mathbf{x}}=F_{\new}(\mathbf{x},t)$ initialized at the current state. For $\tau$, define $\mathbf{y}_\lambda(\tau):=\phi_{\rec}(\tau+\lambda,\tau,\mathbf{x}_{\new}(\tau))$, which is the terminal state obtained by applying $F_{\rec}$ on $[\tau,\tau+\lambda]$ from $\mathbf{x}_{\new}(\tau)$. Assume both flows exist over every candidate insertion interval.

Define the finite insertion objective
\begin{equation}
\begin{aligned}
    J_\lambda(\tau)
    :=\ &
    \mathcal V_{\matM}(\mathbf{x}_{\new}(\tau),\tau)
    +
    \mathcal V_{\matM}(\mathbf{y}_\lambda(\tau),\tau+\lambda)
    \\
    &-
    \mathcal V_{\matM}(\mathbf{x}_{\new}(\tau+\lambda),\tau+\lambda)
    \\
    &+
    \frac12
    \|u_{\rec}(\tau)-u_{\new}(\tau)\|_{\mathbf R}^2.
\end{aligned}
    \label{eq:root_switching_objective}
\end{equation}
The terminal-energy pair is the finite insertion difference in \eqref{eq:finite_insertion_identity}, since $\mathbf{x}_{\new}(\tau+\lambda)=\phi_{\new}(\tau+\lambda,\tau,\mathbf{x}_{\new}(\tau))$. Hence the objective rewards improvement caused by recovery rather than by the deployed flow. Here $\mathbf R=\mathbf R^\top\succeq0$, $\|v\|_{\mathbf R}^2:=v^\top\mathbf Rv$, and $u_{\rec}$ and $u_{\new}$ are the controls inducing $F_{\rec}$ and $F_{\new}$ at the candidate time. Set the quadratic penalty to zero if no such representation is used. The terms balance absolute recoverability, incremental recovery benefit, and handoff abruptness.

\begin{corollary}[CIG consistency]
\label{cor:finite_insertion_objective_consistency}
Assume the hypotheses of Corollary~\ref{cor:infinitesimal_insertion_derivative} and set $\mathbf{x}(\tau)=\mathbf{x}_{\new}(\tau)$. Then \eqref{eq:root_switching_objective} satisfies
\begin{align}
    J_\lambda(\tau)
    ={}&
    \mathcal V_{\matM}(\mathbf{x}_{\new}(\tau),\tau)
    +
    \lambda\G_{\matM}(\mathbf{x}_{\new}(\tau),\tau)
    \nonumber\\
    &+
    \frac12\|u_{\rec}(\tau)-u_{\new}(\tau)\|_{\mathbf R}^2
    +o(\lambda)
    \text{ as }\lambda\downarrow0.
    \label{eq:finite_insertion_objective_expansion}
\end{align}
\end{corollary}

\begin{proof}
The two terminal-energy terms in \eqref{eq:root_switching_objective} are the left-hand side of \eqref{eq:finite_insertion_first_order}. Substitution gives \eqref{eq:finite_insertion_objective_expansion}.
\end{proof}

\begin{definition}[Switching function]
Assume $J_\lambda$ is differentiable and let $\Phi_\lambda(\tau):=\dd J_\lambda(\tau)/\dd\tau$. A zero of $\Phi_\lambda$ is a stationary point of $J_\lambda$ on the interior of the interval for \eqref{eq:root_switching_objective}.
\end{definition}

The derivative of the recovery endpoint in $J_\lambda$ uses the covector $\rho_{\matM}(\mathbf{y}_\lambda(\tau),\tau+\lambda)$. By Corollary~\ref{cor:endpoint_covector}, this covector is computed from the endpoint velocity of the minimizing geodesic for $\mathcal V_{\matM}$. Let $\Psi_{\rec}(b,a;z)$ be the state-transition matrix of the recovery variational dynamics along $s\mapsto\phi_{\rec}(s,a,z)$, where
\begin{equation}
    \frac{\dd}{\dd b}\Psi_{\rec}(b,a;z)=\mathbf A_{\rec}(\phi_{\rec}(b,a,z),b)\Psi_{\rec}(b,a;z)
    \label{eq:recovery_stm}
\end{equation}
with $\Psi_{\rec}(a,a;z)=\mathbf I$. Then, differentiating $\mathbf{y}_\lambda(\tau)=\phi_{\rec}(\tau+\lambda,\tau,\mathbf{x}_{\new}(\tau))$ gives
\begin{align*}
    \frac{\dd \mathbf{y}_\lambda}{\dd\tau}
    &=
    F_{\rec}(\mathbf{y}_\lambda(\tau),\tau+\lambda)
    \nonumber+
    \Psi_{\rec}(\tau+\lambda,\tau;\mathbf{x}_{\new}(\tau))
    \nonumber\\
    &\qquad\cdot
    \bigl(
    F_{\new}(\mathbf{x}_{\new}(\tau),\tau)
    -
    F_{\rec}(\mathbf{x}_{\new}(\tau),\tau)
    \bigr).
\end{align*}
The first term is the terminal-time derivative. The second term is the sum of the initial-time and initial-state derivatives, propagated by the transition matrix \eqref{eq:recovery_stm}. The endpoint obtained by continuing under the deployed mode satisfies $\mathbf{x}_{\new}(\tau+\lambda)=\phi_{\new}(\tau+\lambda,\tau,\mathbf{x}_{\new}(\tau))$, so its total derivative with respect to $\tau$ is $F_{\new}(\mathbf{x}_{\new}(\tau+\lambda),\tau+\lambda)$.

Assume, when the quadratic penalty is retained, that the induced control signals are differentiable functions of $\tau$. Then
\begin{align}
    \Phi_\lambda(\tau)
    &=
    \dot{\mathcal V}_{\matM,\new}
    (\mathbf{x}_{\new}(\tau),\tau)
    \nonumber\\
    &+
    \frac{\partial\mathcal V_{\matM}}{\partial t}
    (\mathbf{y}_\lambda(\tau),\tau+\lambda)
    +
    \rho_{\matM}(\mathbf{y}_\lambda(\tau),\tau+\lambda)^\top
    \frac{\dd \mathbf{y}_\lambda}{\dd\tau}
    \nonumber\\
    &-
    \dot{\mathcal V}_{\matM,\new}
    (\mathbf{x}_{\new}(\tau+\lambda),\tau+\lambda)
    \nonumber\\
    &+
    \bigl(u_{\rec}(\tau)-u_{\new}(\tau)\bigr)^\top
    \mathbf R
    \bigl(\dot u_{\rec}(\tau)-\dot u_{\new}(\tau)\bigr).
    \label{eq:switching_function_derivative}
\end{align}
When $\mathcal V_{\matM}$ is twice differentiable in $\mathbf{x}$ on the prediction segment, let $H_{\matM}(\mathbf{x},t):=\nabla_{\mathbf{x}}^2\mathcal V_{\matM}(\mathbf{x},t)$, and assume $\rho_{\matM}$ and $\Delta F$ are also differentiable in time. Then the derivative of $\G_{\matM}$ along the prediction of the deployed mode is
\begin{align}
    \dot{\G}_{\matM}
    &=
    \left(
    \frac{\partial\rho_{\matM}}{\partial t}
    +
    H_{\matM} F_{\new}
    \right)^\top\Delta F
    \nonumber\\
    &\quad
    +
    \rho_{\matM}^\top
    \left(
    \frac{\partial\Delta F}{\partial t}
    +
    \frac{\partial\Delta F}{\partial \mathbf{x}}F_{\new}
    \right).
    \label{eq:Gdot}
\end{align}
All quantities in \eqref{eq:Gdot} are evaluated at
$(\mathbf{x}_{\new}(\tau),\tau)$. The term $\partial_t\rho_{\matM}$ includes the time variation of $\matM$ and $\mathbf{x}_r$.

The domain for the switching time must exclude reentry into the tube after the deployed trajectory has already lost its certificate. Let
\begin{equation}
    g(\tau)
    :=
    \mathcal V_{\matM}(\mathbf{x}_{\new}(\tau),\tau)-E_{\max}.
    \label{eq:tube_constraint_function}
\end{equation}
and assume $g$ is continuous on the prediction window $[t,t+T]$ with $g(t)\le0$. Define the first certified exit
\begin{equation}
    \tau_{\mathrm{exit}}
    :=
    \inf\{s\in[t,t+T]:g(s)>0\},
    \label{eq:first_certified_exit}
\end{equation}
with $\tau_{\mathrm{exit}}:=t+T$ if the set is empty. The function $g$ is the signed energy residual, so continuity gives $g(s)\le0$ for every $s\in[t,\tau_{\mathrm{exit}}]$. Assuming that the deployed input is admissible, the switching problem over the certified interval is
\begin{equation}
    \tau^\star
    \in
    \argmin_{\tau\in[t,\tau_{\mathrm{exit}}]}J_\lambda(\tau)
    \quad
    \text{s.t.}
    \quad
    g(\tau)\le0.
    \label{eq:constrained_switching_problem}
\end{equation}
The explicit inequality retains the tube feasibility test used by the online solver, while restricting the domain to the first exit certifies the entire deployed segment rather than only the switching endpoint. A later feasible reentry therefore cannot be selected after a prior loss of the certificate. Moreover, if $J_\lambda$ is continuous, the feasible set is nonempty and compact because it contains $t$ and $g\le0$ throughout $[t,\tau_{\mathrm{exit}}]$. Hence \eqref{eq:constrained_switching_problem} attains a global minimum.

\begin{proposition}[First certified exit and latest pre-exit switch]
\label{prop:latest_certified_switch}
Suppose Assumptions~\ref{ass:contracting_recovery_tube} and~\ref{ass:safe_recovery_tube} hold, along with the continuity and initial-feasibility conditions above and
\begin{equation*}
u_{\new}(\mathbf{x}_{\new}(s),s)
\in\mathcal U(\mathbf{x}_{\new}(s),s),
\qquad s\in[t,\tau_{\mathrm{exit}}].
\end{equation*}
Then $g(\tau)\le0$ for every $\tau\in[t,\tau_{\mathrm{exit}}]$. For any switch $\tau$ in this interval, the state and active input satisfy their constraints for every $s\ge t$, and the recovery distance decays exponentially for $s\ge\tau$. If $\tau_{\mathrm{exit}}<t+T$, then $g(\tau_{\mathrm{exit}})=0$, so $\tau_{\mathrm{exit}}$ is the latest certified switching time before the first loss of the recovery certificate. In this case, if $g$ is $C^1$ and
\begin{equation*}
    g'(\tau_{\mathrm{exit}})
    =
    \dot{\mathcal V}_{\matM,\new}
    (\mathbf{x}_{\new}(\tau_{\mathrm{exit}}),\tau_{\mathrm{exit}})
    >0,
\end{equation*}
then $\tau_{\mathrm{exit}}$ is a transversal local exit time and the scalar equation $g(\tau)=0$ has a unique root in a neighborhood of $\tau_{\mathrm{exit}}$.
\end{proposition} 

\begin{proof}
By definition of the infimum, no point in $[t,\tau_{\mathrm{exit}})$ can satisfy $g>0$. If $\tau_{\mathrm{exit}}<t+T$, continuity and $g(t)\le0$ imply $g(\tau_{\mathrm{exit}})=0$. If $g(\tau_{\mathrm{exit}})>0$, then $g>0$ strictly before $\tau_{\mathrm{exit}}$, while if $g(\tau_{\mathrm{exit}})<0$, then no point with $g>0$ can occur arbitrarily close to $\tau_{\mathrm{exit}}$ from the right. Hence $g(s)\le0$ on $[t,\tau]$ for every $\tau\le\tau_{\mathrm{exit}}$. The deployed state therefore remains in the safe recovery tube before the switch, and the assumed condition on the deployed input gives input admissibility there. Theorem~\ref{thm:contractive_recoverability} supplies state safety, admissibility of the recovery input, and exponential convergence after the switch. The derivative identity follows from the chain rule and the definition of $\dot{\mathcal V}_{\matM,\new}$ in \eqref{eq:energy_rate}. When $\tau_{\mathrm{exit}}<t+T$ and $g'(\tau_{\mathrm{exit}})>0$, $g'$ remains positive on a neighborhood of $\tau_{\mathrm{exit}}$. Thus, $g$ is strictly increasing there, which gives a transversal crossing and a locally unique root.
\end{proof}

Proposition~\ref{prop:latest_certified_switch} identifies $\tau_{\mathrm{exit}}$ as the last certified moment in the initial interval within the tube. Subject to its pathwise state and input conditions, the deployed mode may pursue performance until that time, and the switching program is free to place the switch earlier.

\begin{proposition}[Switching-time first-order conditions]
\label{prop:switching_kkt}
Assume $J_\lambda$ and $g$ are continuously differentiable. If $\tau^\star$ locally minimizes \eqref{eq:constrained_switching_problem} and the one-dimensional constraint qualification holds, then there exists $\mu^\star\ge0$ such that
\begin{equation}
\begin{aligned}
    \Phi_\lambda(\tau^\star)+\mu^\star g'(\tau^\star)&=0,
    \qquad
    g(\tau^\star)\le0,
    \\
    \mu^\star&\ge0,
    \qquad
    \mu^\star g(\tau^\star)=0,
\end{aligned}
    \label{eq:KKT_complementarity}
\end{equation}
when $\tau^\star\in(t,\tau_{\mathrm{exit}})$. If $\tau^\star$ lies on the boundary of $[t,\tau_{\mathrm{exit}}]$, the stationarity condition is
\begin{equation*}
    0
    \in
    \Phi_\lambda(\tau^\star)
    +
    \mu^\star g'(\tau^\star)
    +
    N_{[t,\tau_{\mathrm{exit}}]}(\tau^\star),
\end{equation*}
where $N_{[t,\tau_{\mathrm{exit}}]}(\tau^\star)$ denotes the normal cone to $[t,\tau_{\mathrm{exit}}]$ at $\tau^\star$.
\end{proposition}

\begin{proof}
This is the Karush--Kuhn--Tucker theorem applied to the constrained optimization problem \eqref{eq:constrained_switching_problem}. At an interior active constraint, the qualification is $g'(\tau^\star)\ne0$. At an inactive constraint, \eqref{eq:KKT_complementarity} gives $\mu^\star=0$. The normal cone is $\R_{\le0}$ at $t$ and $\R_{\ge0}$ at $\tau_{\mathrm{exit}}$, accounting for solutions at each endpoint. At such a point, stationarity does not require $\Phi_\lambda$ to vanish.
\end{proof}

For a global solution of \eqref{eq:constrained_switching_problem}, interior inactive points satisfying $\Phi_\lambda(\tau)=0$, feasible active-constraint points, and both endpoints of the first certified interval must be compared. A root of the switching function is therefore only a candidate unless this comparison is completed.

\begin{theorem}[Feasible switching implies recoverability]
\label{thm:root_finding_certified_recoverability}
Suppose Assumptions~\ref{ass:contracting_recovery_tube} and~\ref{ass:safe_recovery_tube} hold, $g$ is continuous with $g(t)\le0$, and the deployed input is admissible on $[t,\tau_{\mathrm{exit}}]$, with $\tau_{\mathrm{exit}}$ defined by \eqref{eq:first_certified_exit}. If $\tau^\star$ is feasible for \eqref{eq:constrained_switching_problem}, then, before the switch,
\begin{equation*}
    \mathbf{x}_{\new}(s)\in\mathcal S(s),
    \qquad
    u_{\new}(\mathbf{x}_{\new}(s),s)
    \in\mathcal U(\mathbf{x}_{\new}(s),s),
\end{equation*}
for every $s\in[t,\tau^\star]$. At the switch,
\begin{equation*}
    \mathcal V_{\matM}(\mathbf{x}_{\new}(\tau^\star),\tau^\star)\le E_{\max},
\end{equation*}
and switching to $F_{\rec}$ at $\tau^\star$ gives
\begin{align*}
    d_{\matM}(\mathbf{x}(s),\mathbf{x}_r(s),s)
    \le
    e^{-\gamma(s-\tau^\star)}
    d_{\matM}(\mathbf{x}_{\new}(\tau^\star),
    \mathbf{x}_r(\tau^\star),\tau^\star),
\end{align*}
for $s\ge\tau^\star$.
Moreover, for all $s\ge\tau^\star$,
\begin{equation*}
    \mathbf{x}(s)\in\mathcal S(s),
    \qquad
    u_{\rec}(\mathbf{x}(s),s)\in\mathcal U(\mathbf{x}(s),s).
\end{equation*}
Consequently, the state and active input satisfy their constraints for all $s\ge t$.
\end{theorem}

\begin{proof}
Feasibility gives $\tau^\star\le\tau_{\mathrm{exit}}$. By the definition of the first exit and continuity, $g(s)\le0$ for every $s\in[t,\tau^\star]$, so the deployed trajectory remains in $\T_r\subseteq\mathcal S$ before the switch. Admissibility of the deployed input is assumed on the same interval. At $\tau^\star$, the switching state belongs to $\T_r$. After the switch, both $\mathbf{x}$ and $\mathbf{x}_r$ evolve under $F_{\rec}$, so Theorem~\ref{thm:contractive_recoverability} gives the distance estimate, state safety, and admissibility of the recovery input for all later times.
\end{proof} 

\begin{figure}[!t] 
\centering
\includegraphics[width=0.95\columnwidth]{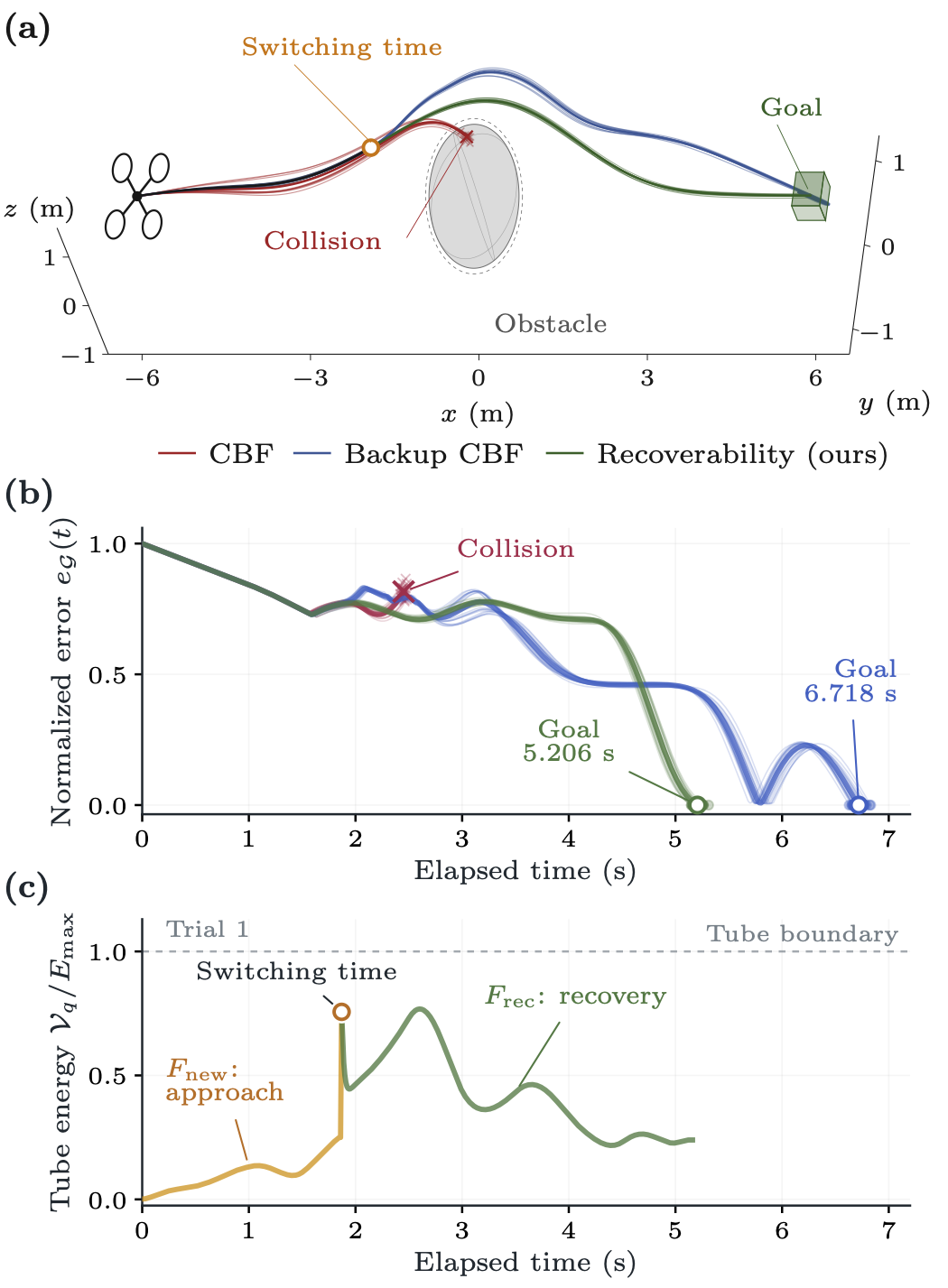}
\caption[Time-optimal quadrotor flight past an obstacle of uncertain size]{Time-optimal quadrotor flight past an obstacle of uncertain size. (a) Ten trials on one side of the obstacle, whose true shape (gray) lies inside the known envelope (dashed). The CBF collides, the backup CBF detours wide, and our framework flies the fast plan until its single certified switch. (b) Normalized goal error, the excess of position and speed over the goal tolerances relative to its initial value, with crosses at collisions and circles at arrivals. Thin curves are trials and thick curves are means aligned by arc length in (a) and by flight duration in (b). (c) Tube energy $\mathcal V_q/E_{\max}$ of trial 1 before (orange) and after (green) the switch, which stays below the tube boundary (dashed).}
\label{fig:quadrotor_scheduling}
\end{figure}

Theorem~\ref{thm:root_finding_certified_recoverability} decouples safety from optimality. Under its pathwise hypotheses, feasibility of the switching time certifies recovery, so a suboptimal feasible solution of \eqref{eq:constrained_switching_problem} may degrade the handoff but never the guarantee. Moreover, the guarantee does not depend on the monitor's prediction of the deployed trajectory being correct. Since the monitor rechecks the measured state against the tube every horizon, Proposition~\ref{prop:tube_preservation} ensures that a switch commanded while the measured state lies in the tube with a nonnegative one-step margin keeps it there until the next check. 

The performance objective belongs to the deployed policy or planner, while the supervisor certifies whether its execution can continue without losing recoverability. Preserving the deployed commands before the switch therefore preserves the intended behavior without asserting that every trajectory inside the tube is performance-optimal. In the single-switch formulation, the recovery mode remains active thereafter, while returning to the deployed mode requires an additional certified re-entry condition.

\subsection{Time-Optimal Quadrotor Flight Under Uncertainty}
\label{sec:quadrotor_scheduling}

Assuming a recovery tube is given a priori, the recoverability framework is now complete. To benchmark it, we consider a time-optimal quadrotor flight motion planning task around an obstacle of uncertain size. This task is dynamically complex, high-performance, and safety-critical, presenting an ideal setting for validating our approach. While the fastest paths are the ones that fly closest to the obstacle, planning for too small an obstacle can lead to collision. 

We benchmark our recoverability framework against a standard CBF~\cite{tayal2024collision} and a backup CBF~\cite{chen2021backup}. We use the nonlinear quadrotor dynamics, parameters, and rotor limits of~\cite{fan2017online}, with state $\xi=(p,v,\theta,\omega)$ comprising inertial position and velocity, attitude in exponential coordinates, and body angular velocity. The vehicle flies from near $(-6,0,0)$~m to a goal box at $(6,0,0)$~m past an ellipsoidal obstacle of random yaw and semi-axes between $0.66$ and $0.80$~m, so it always lies inside a known envelope of radius $0.80$~m. The objective is to minimize the flight time while remaining safe and admissible, with goal-box halfwidths of $(0.25,0.20,0.15)$~m and a speed tolerance of $0.08$~m/s. Across twenty initially feasible trials, we vary the initial state, reference, obstacle, and wind gusts bounded by $0.12$~m/s$^2$, and all methods face identical conditions in each trial without knowing the true obstacle.
\begin{table}[t]
\caption{Comparison of Quadrotor Results.}
\label{tab:quadrotor_obstacle_outcomes}
\centering\footnotesize
\setlength{\tabcolsep}{2.5pt}
\begin{tabular*}{\columnwidth}{@{\extracolsep{\fill}}lcccc@{}}\hline
Method & Success & Time to goal (s) & Path length (m) & Obs. Clear. (m) \\\hline
CBF & $0/20$ & --- & --- & --- \\
Backup CBF & $20/20$ & $6.718\pm0.060$ & $13.401$ & $0.227$ \\
\textbf{Recoverability} & $\mathbf{20/20}$ & $\mathbf{5.206\pm0.045}$ & $\mathbf{12.338}$ & $\mathbf{0.146}$ \\\hline
\multicolumn{2}{@{}l}{Mean paired reduction} & $\mathbf{22.5\%}$ & $\mathbf{7.9\%}$ & $\mathbf{29.6\%}$ \\\hline
\end{tabular*}
\end{table}

In our framework, the deployed mode $F_{\mathrm{new}}$ tracks a time-optimal plan that presupposes an optimistic ball of radius $0.30$~m, smaller than every admissible obstacle. The recovery mode $F_{\mathrm{rec}}$ presupposes the full envelope and follows a conservative bypass that clears it. Both combine feedforward from differential flatness, LQR feedback linearized at hover, and rotor saturation. The provided tube is the sublevel set $\mathcal V_q(\xi,t)=\tfrac12(\xi-\xi_r(t))^\top \mathbf{P}(\xi-\xi_r(t))\le E_{\max}$ of the recovery controller's metric $\mathbf{P}\succ0$ around the simulated recovery trajectory $\xi_r(t)$, with $E_{\max}$ chosen so the tube permits at most $0.08$~m of position deviation. Interval arithmetic over $351\,654$ cells verifies offline that all twenty tubes satisfy the hypotheses of Theorem~\ref{thm:contractive_recoverability} for $t\in[0,14]$~s, with contraction rate $\gamma=0.14$~s$^{-1}$. The framework checks the tube constraint~\eqref{eq:tube_constraint_function} along the deployed prediction and switches at the planned minimizer of~\eqref{eq:root_switching_objective}, with $\lambda=0.10$~s and $\mathbf{R}=0.01\mathbf I_4$, or at first contact with the tube boundary, whichever occurs first. 

The backup CBF uses the same deployed and recovery modes, metric, and envelope, together with a $6$-s horizon and a terminal ellipsoid at hover at the goal. Its published quadratic program blends the commands of the two modes instead of switching between them, so the comparison isolates filtering from switching. The standard CBF runs its published quadratic program on the deployed mode alone, enforcing avoidance of the optimistic $0.30$~m ball.

Figure~\ref{fig:quadrotor_scheduling} summarizes our results. The standard CBF collides in all twenty trials, because the set it renders invariant is not the true safe set. The backup CBF reaches the goal in every trial but pays for its filtering approach in terms of performance. Our framework is also safe in every trial and arrives first in every pair, even though its switching objective~\eqref{eq:root_switching_objective} does not reward arrival time. Table~\ref{tab:quadrotor_obstacle_outcomes} lists mean $\pm$ sample SD arrival times, mean path lengths, and minimum clearances. In mean paired comparisons, recoverability reaches the goal in $22.5\%$ less time along a $7.9\%$ shorter path than the backup CBF. In $7$ trials the measured state reached the tube boundary before the planned switch, and switching there recovered the vehicle every time. No recoverability trial left its tube, even under gusts above the sufficient robustness limit of Theorem~\ref{thm:robust_recoverability}, as Fig.~\ref{fig:quadrotor_scheduling}(c) illustrates for one trial.

\section{Synthesis of Recoverability Certificates}
\label{sec:certificate_synthesis}

In the previous quadrotor example, the recovery tube was fixed in advance. However, the tube defines where the deployed mode may operate and still be recovered, so a larger tube directly expands the certified operating envelope. To this end, this section presents a synthesis framework for maximal recoverability certificates. We pose an ideal problem that searches jointly over the recovery policy, metric, reference, and radius for the largest certified tube, and show that every feasible point already carries the recoverability guarantee. We then construct a restricted transport family whose distance stays in closed form, so the online test remains a single evaluation, and grow it by sequential convex programming toward the viability kernel.

Let the policies $\mathbf{k}_{\new}$ and $\mathbf{k}_{\rec}$ induce the fields $F_{\new}$ and $F_{\rec}$, respectively, with $u_{\new}=\mathbf{k}_{\new}(\mathbf{x},t)$ and $u_{\rec}=\mathbf{k}_{\rec}(\mathbf{x},t)$, where $\mathbf{k}_{\new}$ is fixed and $\mathbf{k}_{\rec}$ is synthesized.
For a metric $\matM$, reference $\mathbf{x}_r$, rate $\gamma$, and radius $r$, define
\begin{equation*}
    \T_r(t)
    :=
    \{\mathbf{x}:d_{\matM}(\mathbf{x},\mathbf{x}_r(t),t)\le r(t)\}.
\end{equation*}
For $\mathbf{x}\in\T_r(t)$, let $\xi_s$ be the state reached at time $t+s$ by the deployed trajectory $\mathbf{x}_{\new}$ started from $\mathbf{x}$ at time $t$, let $u_s^{\new}:=\mathbf{k}_{\new}(\xi_s,t+s)$, and define
\begin{equation}
\begin{aligned}
    \tau_{\new}(\mathbf{x},t)
    :=
    \max\bigl(\{ \tau\in[0,H]:\ &
    \xi_s\in\T_r(t+s),\\[-1mm]
    &u_s^{\new}\in\mathcal U(\xi_s,t+s),\\[-1mm]
    &\forall s\in[0,\tau]\}\cup\{0\}\bigr).
\end{aligned}
    \label{eq:certified_deployed_horizon}
\end{equation}
Assume the maximum in \eqref{eq:certified_deployed_horizon} exists, and zero permits an instant switch. Under \eqref{eq:synthesis_safety}, $\xi_s\in\T_r(t+s)$ implies $\xi_s\in\mathcal S(t+s)$.

The synthesis objective combines tube volume and admissible horizon \eqref{eq:certified_deployed_horizon}. With decision tuple $\mathcal C:=(\mathbf{k}_{\rec},\matM,\mathbf{x}_r,r,\gamma)$, the ideal recoverability synthesis problem is
\begin{subequations}
\label{eq:ideal_recoverability_synthesis}
\begin{align}
    \sup_{\mathcal C} 
    &
    \begin{aligned}[t]
    &
    \int_0^{T_f} \! \!
    \operatorname{vol}(\T_r(t))\dd t
    +
    \alpha\!
    \int_0^{T_f} \! \!
    \int_{\T_r(t)} \! \! \! \!
    \tau_{\new}(\mathbf{x},t)\eta(\mathbf{x},t)
    \dd \mathbf{x}\dd t
    \end{aligned}
    \\
    \text{s.t.}\quad
    &
    \dot{\mathbf{x}}_r(t)=F_{\rec}(\mathbf{x}_r(t),t),
    \label{eq:synthesis_reference}
    \\
    &
    \underline m \mathbf I
    \preceq
    \matM(\mathbf{x},t)
    \preceq
    \overline m \mathbf I,
    \notag
    \\
    &
    \dot{\matM}_{\rec}
    +
    \mathbf A_{\rec}^\top \matM
    +
    \matM \mathbf A_{\rec}
    \preceq
    -2\gamma(t)\matM,
    \label{eq:synthesis_contraction}
    \\
    &
    \dot r(t)\ge -\gamma(t)r(t),
    \
    r(t)>0,\
    \gamma(t)\ge\gamma_{\min}>0,
    \label{eq:synthesis_radius}
    \\
    &
    \T_r(t)\subseteq\mathcal S(t),
    \label{eq:synthesis_safety}
    \\
    &
    \mathbf{k}_{\rec}(\mathbf{x},t)\in\mathcal U(\mathbf{x},t),
    \label{eq:synthesis_input}
\end{align}
\end{subequations}
Constraint~\eqref{eq:synthesis_contraction} is imposed on an open neighborhood of $\T_r(t)$, and the remaining pointwise constraints hold for every $\mathbf{x}\in\T_r(t)$. For Theorem~\ref{thm:synthesis_feasibility_guarantee}, Assumptions~\ref{ass:contracting_recovery_tube} and \ref{ass:safe_recovery_tube} are applied to the state-time tube $\{(\mathbf{x},t):\mathbf{x}\in\T_r(t)\}$, and Definition~\ref{def:recovery_contraction_metric} uses rate $\gamma(t)$. Here $\eta(\mathbf{x},t)\ge0$ is a weighting function, and $\alpha\ge0$ weights the second integral relative to the first. Intuitively, the first integral rewards a large tube, the set of states from which recovery is certified, while the second rewards tubes in which the deployed mode can keep running longer before a switch is required, so $\alpha$ trades the size of the certified envelope against how long aggressive operation can continue inside it.

\begin{theorem}[Feasible synthesis]
\label{thm:synthesis_feasibility_guarantee}
Let $(\mathbf{k}_{\rec},\matM,\mathbf{x}_r,r,\gamma)$ be feasible for \eqref{eq:ideal_recoverability_synthesis}, with $r$ continuously differentiable and $\gamma$ continuous. For any $\mathbf{x}\in\T_r(t)$, suppose the deployed trajectory $\mathbf{x}_{\new}$ started from $\mathbf{x}$ at time $t$ exists on $[t,t+\tau]$ and $\tau\le\tau_{\new}(\mathbf{x},t)$. If the system follows $F_{\new}$ on $[t,t+\tau]$ and then switches to $F_{\rec}$, the trajectory remains in $\mathcal S(s)$ and its active input is admissible for all $s\ge t$. Let $L(s):=d_{\matM}(\mathbf{x}(s),\mathbf{x}_r(s),s)$. After the switch,
\begin{align*}
    L(s)
    &\le
    \exp\!\left(-\int_{t+\tau}^{s}\gamma(\sigma)\dd\sigma\right)
    L(t+\tau),
    \\
    L(s)
    &\le r(s),
    \qquad s\ge t+\tau.
\end{align*}
\end{theorem} 

\begin{proof}
The definition of $\tau_{\new}$ and \eqref{eq:synthesis_safety} give state and input admissibility on $[t,t+\tau]$ and place the switching state in $\T_r(t+\tau)$. Lemma~\ref{lem:path_contraction}, applied with the rate $\gamma(s)$ of~\eqref{eq:synthesis_contraction}, gives $D^+L(s)\le-\gamma(s)L(s)$ while $L(s)\le r(s)$. Hence
\begin{equation*}
    L(s)
    \le
    \exp\!\left(-\int_{t+\tau}^{s}\gamma(\sigma)\dd\sigma\right)
    L(t+\tau).
\end{equation*}
Moreover, \eqref{eq:synthesis_radius} gives
\begin{equation*}
    r(s)
    \ge
    \exp\!\left(-\int_{t+\tau}^{s}\gamma(\sigma)\dd\sigma\right)
    r(t+\tau).
\end{equation*}
Since $L(t+\tau)\le r(t+\tau)$, the comparison lemma applied up to a first exit gives $L(s)\le r(s)$ throughout the maximal recovery interval. Assumption~\ref{ass:safe_recovery_tube} makes this interval unbounded, and \eqref{eq:synthesis_safety}--\eqref{eq:synthesis_input} give safety and input admissibility.
\end{proof}

Because the guarantee of Theorem~\ref{thm:synthesis_feasibility_guarantee} holds at every feasible point of \eqref{eq:ideal_recoverability_synthesis} independently of the objective, it suffices to construct feasible points, as we do in the following section.

\subsection{Transport Certificates With Closed-Form Distance}
\label{subsec:contractive_transport}

Since \eqref{eq:ideal_recoverability_synthesis} is intractable in general, we use Theorem~\ref{thm:synthesis_feasibility_guarantee} to develop a constructive approach that searches a restricted certificate family through a tractable optimization. Its parameterization transports the problem to latent coordinates in which the recovery dynamics are linear and the certificate objects are available in closed form. We restrict \eqref{eq:ideal_recoverability_synthesis} to an autonomous equilibrium, a fixed reference $\mathbf{x}_r$, and a fixed safe set $\mathcal S$. These restrictions define a tractable certificate family rather than a requirement of recoverability, since Theorems~\ref{thm:contractive_recoverability}--\ref{thm:root_finding_certified_recoverability} hold for any recovery contraction metric. Let $\Phi:\X\to\R^n$ be a $C^2$ diffeomorphism defining the latent coordinates $\mathbf{z}=\Phi(\mathbf{x})$, let $\mathbf W\succ0$, and define
\begin{align}
    q_{\Phi,\mathbf W}(\mathbf{x})
    &:=
    \Phi(\mathbf{x})^\top
    \mathbf W^{-1}
    \Phi(\mathbf{x}),
    \nonumber\\
    \matM_{\Phi,\mathbf W}(\mathbf{x})
    &:=
    D\Phi(\mathbf{x})^\top
    \mathbf W^{-1}
    D\Phi(\mathbf{x}),
    \label{eq:transport_metric}
\end{align}
as well as the transported tube of radius one,
\begin{equation*}
    \T_{\Phi,\mathbf W}
    :=
    \{\mathbf{x}:q_{\Phi,\mathbf W}(\mathbf{x})\le1\}.
\end{equation*}
The matrix $\mathbf W$ defines an ellipsoid in the latent coordinates, and $\Phi^{-1}$ maps that ellipsoid to the physical recovery tube. With $\Phi$ the identity, the family reduces to the ellipsoidal certificates of classical linear matrix inequality synthesis~\cite{boyd1994linear}. Let the latent recovery dynamics be $\dot{\mathbf{z}}=\mathbf A\mathbf{z}$ with $\mathbf A=\mathbf Y\mathbf W^{-1}$, a factorization that renders the latent contraction condition linear in $(\mathbf Y,\mathbf W)$. The desired pullback field is
\begin{equation}
    F^d_{\rec,\Phi}(\mathbf{x})
    =
    D\Phi(\mathbf{x})^{-1}
    \mathbf Y\mathbf W^{-1}\Phi(\mathbf{x}).
    \label{eq:transport_recovery_field}
\end{equation}
Requiring $\Phi(\mathbf{x}_r)=0$ places the reference at the latent origin, and requiring $\det D\Phi\equiv1$ makes $\Phi$ volume preserving, so the volume of the tube depends on $\mathbf W$ alone. In this restricted setting, we require the recovery policy $\mathbf{k}_\Phi$ to make the system linear in the latent coordinates, which a fully actuated plant permits exactly (see Section~\ref{sec:kinematic_car_example}) and other plants permit up to a bounded mismatch.

\begin{proposition}[Transport certificate]
\label{prop:transport_certificate}
Assume that $\mathcal S$ is closed and that $\mathbf{x}_r\in\operatorname{int}\mathcal S$. Let $\Phi$ be a $C^2$ diffeomorphism satisfying $\Phi(\mathbf{x}_r)=0$ and $\det D \Phi\equiv1$. Then
\begin{equation}
    d_{\matM_{\Phi,\mathbf W}}^2(\mathbf{x},\mathbf{x}_r)
    =
    q_{\Phi,\mathbf W}(\mathbf{x}).
    \label{eq:transport_exact_distance}
\end{equation}
Moreover,
\begin{equation}
    \operatorname{vol}(\T_{\Phi,\mathbf W})
    =
    \omega_n\sqrt{\det \mathbf W},
    \label{eq:transport_volume}
\end{equation}
where $\omega_n$ is the volume of the unit ball in $\R^n$. If, in addition, $\mathbf{k}_\Phi$ induces $F^d_{\rec,\Phi}$ on an open neighborhood of $\T_{\Phi,\mathbf W}$ and
\begin{align*} 
    &\mathbf W\succeq\epsilon\mathbf I,
    \qquad
    \mathbf Y+\mathbf Y^\top\preceq-2\gamma\mathbf W,
    \qquad
    \inf_{\mathbf{y}\in\partial\mathcal S}q_{\Phi,\mathbf W}(\mathbf{y})\ge1,\\
    &\mathbf{k}_\Phi(\mathbf{x})\in\mathcal U(\mathbf{x})
    \quad
    \forall\mathbf{x}\in\T_{\Phi,\mathbf W},
\end{align*}
then $(\mathbf{k}_\Phi,\matM_{\Phi,\mathbf W},\mathbf{x}_r,1,\gamma)$ is feasible for \eqref{eq:ideal_recoverability_synthesis}, and recovery after a switch at time $t_s$ from any state in $\T_{\Phi,\mathbf W}$ satisfies
\begin{equation*}
    d_{\matM_{\Phi,\mathbf W}}(\mathbf{x}(t),\mathbf{x}_r)
    \le
    e^{-\gamma(t-t_s)}
    d_{\matM_{\Phi,\mathbf W}}(\mathbf{x}(t_s),\mathbf{x}_r),
    \quad
    \forall t\ge t_s,
\end{equation*}
while remaining in $\mathcal S$ and satisfying the input constraints.
\end{proposition}

\begin{proof}
For any path $\bfc$ from $\mathbf{x}_r$ to $\mathbf{x}$, the change of variables $\mathbf{z}=\Phi(\bfc)$ gives
\begin{equation*}
    \dot{\bfc}^{\top}\matM_{\Phi,\mathbf W}(\bfc)\dot{\bfc}
    =
    \dot{\mathbf{z}}^{\top}\mathbf W^{-1}\dot{\mathbf{z}}.
\end{equation*}
The map $\Phi$ bijects physical and latent paths, where the straight path minimizes length, proving \eqref{eq:transport_exact_distance}. Its unit Jacobian gives \eqref{eq:transport_volume}. The tube is compact, so its metric has uniform positive bounds. Its strict interior is connected, contains $\mathbf{x}_r$, and is disjoint from $\partial\mathcal S$ because $\inf_{\partial\mathcal S}q_{\Phi,\mathbf W}\ge1$. Taking closures gives $\T_{\Phi,\mathbf W}\subseteq\mathcal S$.

Along \eqref{eq:transport_recovery_field}, $\dot{\mathbf{z}}=\mathbf Y\mathbf W^{-1}\mathbf{z}$, and every infinitesimal deviation $\delta\mathbf z$ satisfies
\begin{align*}
    \frac{\dd}{\dd t}
    \delta\mathbf z^\top\mathbf W^{-1}\delta\mathbf z
    &=
    \delta\mathbf z^\top
    \mathbf W^{-1}(\mathbf Y+\mathbf Y^\top)\mathbf W^{-1}
    \delta\mathbf z
    \\
    &\le
    -2\gamma
    \delta\mathbf z^\top\mathbf W^{-1}\delta\mathbf z,
\end{align*}
so the pullback metric satisfies \eqref{eq:synthesis_contraction}. Exact realization establishes feasibility of the ideal problem with $r\equiv1$.
\end{proof}

Proposition~\ref{prop:transport_certificate} reduces the restricted certificate family to $(\Phi,\mathbf W,\mathbf Y)$. Distance is quadratic, volume is explicit in $\mathbf W$, and contraction is a matrix inequality in $(\mathbf Y,\mathbf W)$. The updates below target the $\alpha=0$ objective while preserving certification. We additionally note that if $\mathbf{k}_\Phi$ realizes $F^d_{\rec,\Phi}$ only up to a mismatch $d_\Phi$ with $\|d_\Phi\|_{\matM_{\Phi,\mathbf W}}\le\bar d_\Phi\le\gamma$ on $\T_{\Phi,\mathbf W}$, and the perturbed recovery solutions are forward complete, Theorem~\ref{thm:robust_recoverability} applies with $r=1$, so $\T_{\Phi,\mathbf W}$ remains forward invariant and
\begin{align}
    d_{\matM_{\Phi,\mathbf W}}(\mathbf{x}(t),\mathbf{x}_r)
    &\le e^{-\gamma(t-t_s)}
    d_{\matM_{\Phi,\mathbf W}}(\mathbf{x}(t_s),\mathbf{x}_r)\notag\\
    &\quad+\frac{\bar d_\Phi}{\gamma}
    (1-e^{-\gamma(t-t_s)}).
    \label{eq:transport_robust_bound}
\end{align}

\subsection{Certified Tube Growth via Sequential Convex Program}
\label{subsec:sequential_convex_transport}

While the previous section showed that any transport $(\Phi,\mathbf W,\mathbf Y)$ satisfying Proposition~\ref{prop:transport_certificate} is a valid certificate with a closed-form tube, this section searches that family for the largest such tube. Enlarging the tube means bending $\Phi$ and inflating $\mathbf W$. Volume and contraction remain convex in $(\mathbf W,\mathbf Y)$, but keeping the bent tube inside the safe set and realizable by the recovery policy depends nonlinearly on $\Phi$. We therefore grow the tube in small steps, as in sequential convex trajectory optimization. Each step solves a convex problem that enforces safety only at sampled boundary points, and an offline check on the entire boundary then accepts or rejects it, as summarized in Algorithm~\ref{alg:transport_synthesis}.

We parameterize $\Phi$ as a composition of volume-preserving shears applied to a seed transport $\Phi^0$, adding one shear per step. Each shear shifts one block of latent coordinates by a weighted sum of fixed basis functions $\psi_j$ of the other, with the weights $\vartheta$ as decision variables. For a split $\mathbf{z}=(\mathbf{a},\mathbf{b})\in\R^p\times\R^{n-p}$, which may change between steps through a coordinate permutation, define
\begin{equation}
\begin{aligned}
    \mathcal L_\vartheta(\mathbf{a},\mathbf{b})
    &:=(\mathbf{a},\mathbf{b}+\Psi_\vartheta(\mathbf{a})),
    \\
    \Psi_\vartheta(\mathbf{a})
    &:=\sum_{j=1}^m\vartheta_j\psi_j(\mathbf{a}),
    \qquad \psi_j(0)=0.
\end{aligned}
    \label{eq:coupling_layer}
\end{equation}
Because $\mathbf{b}$ is shifted by a function of $\mathbf{a}$ alone, the Jacobian of $\mathcal L_\vartheta$ is block triangular with identity diagonal blocks. For every $\vartheta$, the shear therefore has unit Jacobian determinant, is inverted in closed form by $(\mathbf{a},\mathbf{b})\mapsto(\mathbf{a},\mathbf{b}-\Psi_\vartheta(\mathbf{a}))$, and fixes the latent origin because $\psi_j(0)=0$. Composing shears thus bends the tube while preserving $\Phi(\mathbf{x}_r)=0$ and $\det D\Phi\equiv1$, and alternating the split lets successive shears bend every coordinate. Given an accepted transport $\Phi^k$, the candidate update is $\widehat\Phi^{k+1}:=\mathcal L_\vartheta\circ\Phi^k$.

Safety is imposed at stored boundary points $\mathcal A_k=\{\mathbf{y}_i\}_{i=1}^{N_k}\subset\partial\mathcal S$ through $g_i^k(\vartheta,\mathbf W):=\mathcal L_\vartheta(\mathbf{z}_i^k)^\top\mathbf W^{-1}\mathcal L_\vartheta(\mathbf{z}_i^k)$, where $\mathbf{z}_i^k=\Phi^k(\mathbf{y}_i)$. The matrix fractional function is jointly convex and $\mathcal L_\vartheta(\mathbf{z}_i^k)$ is affine in $\vartheta$, so the first-order Taylor approximation $\ell_i^k$ of $g_i^k$ at $(0,\mathbf W^k)$, taken with the Frobenius inner product in $\mathbf W$, is a global lower bound, and $\ell_i^k\ge1$ is a sufficient affine boundary constraint. The $k$th subproblem is
\begin{subequations}
\label{eq:sequential_transport_subproblem}
\begin{align}
    \max_{\vartheta,\mathbf W,\mathbf Y,\kappa}\quad
    &
    \frac12\log\det\mathbf W
    +
    c_\kappa\kappa
    -
    c_\vartheta\|\vartheta\|_2^2
    \label{eq:sct_objective}
    \\
    \text{s.t.}\quad
    &
    \epsilon\mathbf I\preceq\mathbf W\preceq\overline w\mathbf I,
    \label{eq:sct_positive}
    \\
    &
    \mathbf Y+\mathbf Y^\top
    \preceq
    -2\gamma\mathbf W,
    \label{eq:sct_contraction}
    \\
    &
    \ell_i^k(\vartheta,\mathbf W)
    \ge
    1+\kappa,
    \quad
    i=1,\ldots,N_k,
    \label{eq:sct_cuts}
    \\
    &
    0\le\kappa\le\overline\kappa,
    \
    \|\vartheta\|_2\le\nu_k,
    \
    \|\mathbf Y\|_F\le\overline y,
    \label{eq:sct_trust}
    \\
    &
    \log\det\mathbf W
    \ge
    \log\det\mathbf W^k.
    \label{eq:sct_monotone}
\end{align}
\end{subequations}
It maximizes tube volume through $\log\det\mathbf W$, rewards a boundary margin $\kappa$, and penalizes large steps, subject to bounds on $\mathbf W$, the contraction inequality, the sampled boundary constraints, a trust region of radius $\nu_k$, and no loss of volume, with $c_\kappa,c_\vartheta\ge0$. Choosing $\overline w$ and $\overline y$ so that $\mathbf W^0\preceq\overline w\mathbf I$ and $\|\mathbf Y^0\|_F\le\overline y$ makes the feasible set compact, so the subproblem attains its maximum whenever it is nonempty.

Because the subproblem only sees sampled points, a candidate is accepted only if its boundary margin
\begin{equation}
    D_{\widehat\Phi^{k+1},\mathbf W^{k+1}}
    :=
    \inf_{\mathbf{y}\in\partial\mathcal S}
    q_{\widehat\Phi^{k+1},\mathbf W^{k+1}}(\mathbf{y})
    \label{eq:true_boundary_oracle}
\end{equation}
is certified to be at least one and its input and realization conditions are certified. Otherwise the current certificate is kept, any violating boundary point is added to $\mathcal A_k$, and the trust radius is reduced.

\begin{remark}[Polynomial boundary verification]
\label{rem:polynomial_boundary_oracle}
If $\partial\mathcal S$ is a finite union of polynomial curve segments $\mathbf y(s)$, $s\in[0,1]$, and $\Phi^k$ and the basis functions $\psi_j$ are polynomial, then $q_{\widehat\Phi^{k+1},\mathbf W^{k+1}}(\mathbf y(s))$ is a univariate polynomial on each segment. Interval or Bernstein subdivision with outward rounding then yields certified lower bounds on~\eqref{eq:true_boundary_oracle} that converge to its value, so the check terminates when the margin exceeds one.
\end{remark}

\begin{theorem}[Feasible transport updates]
\label{thm:certified_sequential_transport}
Problem~\eqref{eq:sequential_transport_subproblem} is convex, and at every accepted iterate of Algorithm~\ref{alg:transport_synthesis} it is nonempty because the null update $(\vartheta,\mathbf W,\mathbf Y,\kappa)=(0,\mathbf W^k,\mathbf Y^k,0)$ is feasible. If the seed $(\Phi^0,\mathbf W^0)$ satisfies Proposition~\ref{prop:transport_certificate}, every exactly realized accepted iterate defines
\begin{equation}
    \mathcal C^k
    :=
    (\mathbf{k}_{\Phi^k},\matM_{\Phi^k,\mathbf W^k},\mathbf{x}_r,1,\gamma)
    \label{eq:accepted_certificate_tuple}
\end{equation}
which is feasible for \eqref{eq:ideal_recoverability_synthesis}, and robustly realized iterates satisfy \eqref{eq:transport_robust_bound}. Moreover, every accepted update $k\to k+1$ satisfies
\begin{equation}
    \operatorname{vol}(\T_{\Phi^{k+1},\mathbf W^{k+1}})
    \ge
    \operatorname{vol}(\T_{\Phi^k,\mathbf W^k}).
    \label{eq:accepted_volume_monotonicity}
\end{equation}
Rejected updates retain the current feasible certificate, so the accepted sequence is recursively feasible. If $\operatorname{vol}(\mathcal S)<\infty$, the accepted tube volumes converge to a finite limit.
\end{theorem}

\begin{proof}
The objective of~\eqref{eq:sequential_transport_subproblem} is concave in $(\vartheta,\mathbf W,\kappa)$. Constraints \eqref{eq:sct_positive}--\eqref{eq:sct_trust} are convex, and \eqref{eq:sct_monotone} is a superlevel set of concave $\log\det\mathbf W$. Hence \eqref{eq:sequential_transport_subproblem} is convex. At an accepted iterate, $\ell_i^k(0,\mathbf W^k)=g_i^k(0,\mathbf W^k)\ge1$ for every stored boundary point. Together with the accepted contraction condition, the bounds chosen in \eqref{eq:sct_positive} and \eqref{eq:sct_trust}, and equality in \eqref{eq:sct_monotone}, this proves feasibility of $(0,\mathbf W^k,\mathbf Y^k,0)$. The same tuple reproduces the current transport and therefore passes every global feasibility check already verified at iteration $k$.

Each shear is a diffeomorphism fixing the origin with unit Jacobian determinant. At an accepted iterate, \eqref{eq:sct_positive}, \eqref{eq:sct_contraction}, the certified boundary margin~\eqref{eq:true_boundary_oracle}, and the input and realization hypotheses imply Proposition~\ref{prop:transport_certificate}. Equations~\eqref{eq:transport_volume} and \eqref{eq:sct_monotone} give \eqref{eq:accepted_volume_monotonicity}. If $\operatorname{vol}(\mathcal S)<\infty$, the volumes are bounded above by $\operatorname{vol}(\mathcal S)$, so the monotone scalar sequence converges.
\end{proof}

\begin{algorithm}[t]
\caption{Sequential Transport Synthesis}
\label{alg:transport_synthesis}
\algsetup{indent=0.9em}
\begin{algorithmic}[1]
\REQUIRE Seed $(\Phi^0,\mathbf W^0,\mathbf Y^0)$ satisfying Proposition~\ref{prop:transport_certificate}, boundary samples $\mathcal A_0$, trust radius $\nu_0$
\REPEAT
    \STATE Solve \eqref{eq:sequential_transport_subproblem} and form $\widehat\Phi^{k+1}=\mathcal L_\vartheta\circ\Phi^k$
    \IF{The candidate tube is certified to lie inside $\mathcal S$ by \eqref{eq:true_boundary_oracle} and satisfies input and realization conditions}
        \STATE Accept $(\Phi^{k+1},\mathbf W^{k+1},\mathbf Y^{k+1})\gets(\widehat\Phi^{k+1},\mathbf W,\mathbf Y)$ and set $k\gets k+1$
    \ELSE
        \STATE Keep the current certificate, add any violating boundary point to $\mathcal A_k$, and reduce $\nu_k$
    \ENDIF
\UNTIL{The accepted volume stops increasing}
\end{algorithmic}
\end{algorithm}

\subsection{Evaluating Recoverability Against Exact Viability}
\label{sec:kinematic_car_example}

With synthesis in hand, the recovery tube no longer has to be given a priori. We now instantiate the transport construction of Section~\ref{subsec:contractive_transport} on a planar error model whose metric, geodesics, recovery flow, and comparison sets are available in closed form so every comparison is exact. The example measures the synthesized tube against the exact viability kernel and the backup-CBF set, verifies the switching optimum of Section~\ref{sec:root_finding_switching} exactly, and shows that a closed-form certificate need not be an ellipsoid like the quadrotor's.

\begin{figure*}
    \centering
\includegraphics[width=0.9\textwidth]{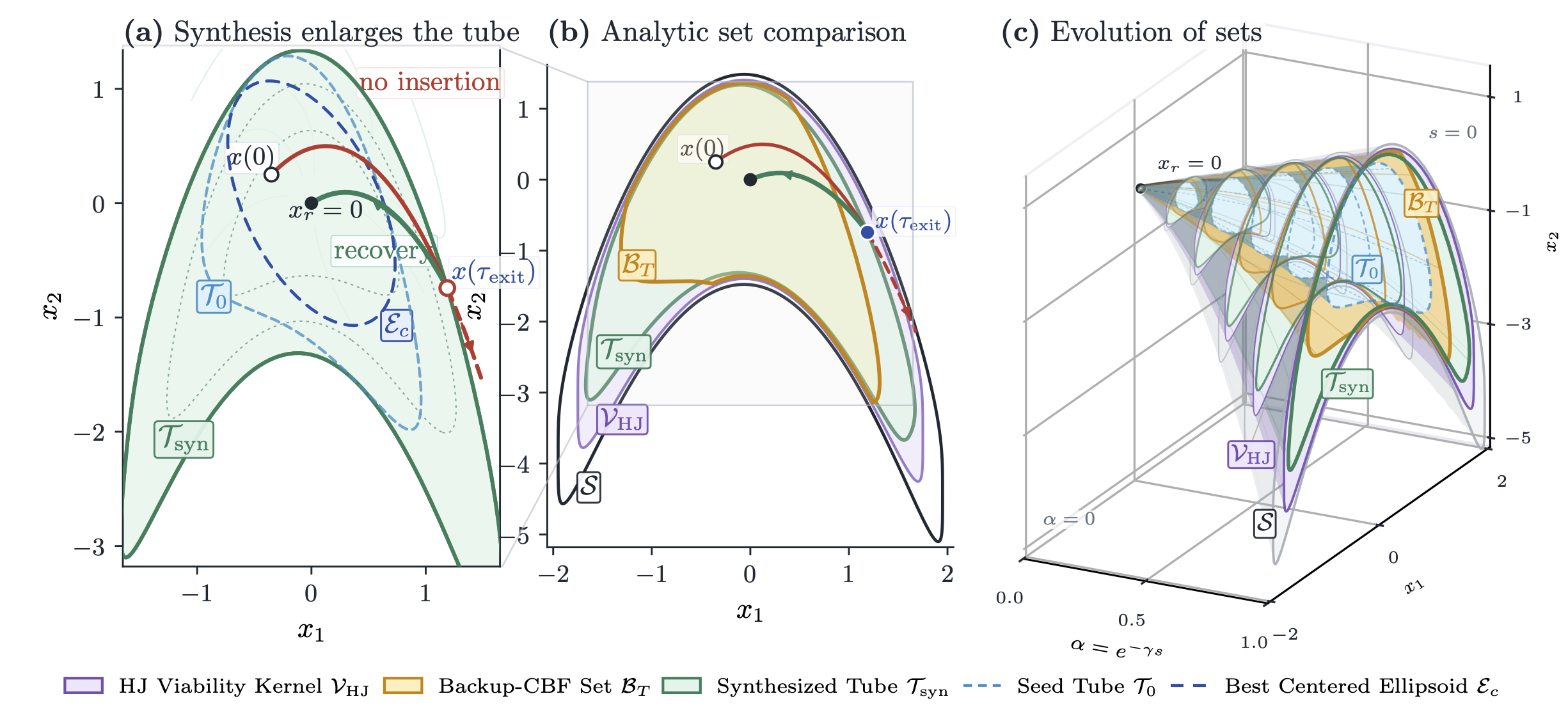}
    \caption{Planar recoverability example.
(a) The synthesized certificate $\T_{\rm syn}$ has greater area than the seed tube $\T_0$ while preserving a closed-form geodesic distance. The dark dashed curve is the largest centered $\mathbf P$-ellipsoid $\mathcal E_c$, the red curve is the deployed no-insertion trajectory, and the green curve is the recovery trajectory after the latest certified switch $\tau_{\rm exit}$ on the tube boundary, which contracts back to $\mathbf{x}_r=0$.
(b) Analytic set comparison for the same planar system, including the state-constraint set $\mathcal S$, the HJ viability kernel $\mathcal V_{\rm HJ}\subseteq\mathcal S$, the finite-horizon backup-CBF rollout set $\mathcal B_T$, and $\T_{\rm syn}$.
(c) Evolution of the set boundaries under the corresponding recovery flows, parameterized by $\alpha=e^{-\gamma s}$. The right cross-section corresponds to $s=0$, and all boundaries converge to $\mathbf{x}_r=0$ as $s\to\infty$.}
    \label{fig:planar_car_recoverability}
\end{figure*}

Let $\mathbf{x}=[x_1,x_2]^\top$ with nonlinear coordinate change
\begin{equation*}
    \mathbf{z}
    =
    \varphi(\mathbf{x})
    :=
    \begin{bmatrix}
        x_1\\
        x_2+\beta x_1^2
    \end{bmatrix},
    \qquad
    \mathbf J(\mathbf{x})
    :=
    \frac{\partial\varphi}{\partial \mathbf{x}}
    =
    \begin{bmatrix}
        1 & 0\\
        2\beta x_1 & 1
    \end{bmatrix}.
\end{equation*}
Since $\det\mathbf J\equiv1$, $\varphi$ is a global, area-preserving diffeomorphism with inverse $\varphi^{-1}(z_1,z_2)=(z_1,z_2-\beta z_1^2)$. The map $\varphi$ is itself a single shear of the form~\eqref{eq:coupling_layer}, with $(\mathbf a,\mathbf b)=(x_1,x_2)$, $\psi_1(\mathbf a)=\mathbf a^2$, and $\vartheta_1=\beta$, so $\varphi$ fixes the origin and is a valid transport map. The coordinates $\mathbf z=\varphi(\mathbf x)$ define the plant and its constraints below, and $\varphi$ also serves as the seed transport $\Phi^0$ of Algorithm~\ref{alg:transport_synthesis}. The inverse map bends horizontal lines in the latent coordinates into parabolas, which is why every set in Fig.~\ref{fig:planar_car_recoverability} is arched.

Then, consider the controlled plant
\begin{equation}
    \dot{\mathbf x}
    =
    \mathbf J(\mathbf x)^{-1}
    \bigl(8\varphi(\mathbf x)+u\bigr),
    \qquad
    u\in\mathcal U,
    \label{eq:planar_controlled_plant}
\end{equation}
whose latent dynamics $\dot{\mathbf z}=8\mathbf z+u$ are an unstable drift that the input must cancel. The state constraint and the input bound $\mathcal U:=8K$ are rounded boxes in the latent coordinates,
\begin{align*}
\mathcal S
&=
\varphi^{-1}\!\left(
\left\{\mathbf z:
\left|\frac{z_1}{1.95}\right|^{4.2}
+
\left|\frac{z_2+0.14z_1}{1.48}\right|^{4.2}
\le1
\right\}
\right),
\\
K
&:=
\left\{\mathbf z:
\left|\frac{z_1}{1.75}\right|^{4.2}
+
\left|\frac{z_2+0.14z_1}{1.40}\right|^{4.2}
\le1
\right\}.
\end{align*}
An admissible input holds every point of $K$ in place through $u=-8\mathbf z$, while outside $K$ the drift outruns every input in $8K$ and carries the state out of the bounded set $\varphi(\mathcal S)$. Hence $\mathcal V_{\rm HJ}:=\varphi^{-1}(K)$ is the exact infinite-horizon HJ viability kernel~\cite{mitchell2005time,bansal2017hamilton}.

Recovery is certified around the reference $\mathbf{x}_r=0$ with the pullback metric
\begin{equation*}
    \mathbf P
    =
    \begin{bmatrix}
        1.60 & 0.55\\
        0.55 & 0.75
    \end{bmatrix}
    \succ0,
    \qquad
    \matM(\mathbf{x})=\mathbf J(\mathbf{x})^\top\mathbf P\mathbf J(\mathbf{x}).
\end{equation*}
The seed tube $\T_0:=\{\mathbf{x}:d_{\matM}(\mathbf{x},0)\le r\}$ is the transported tube $\T_{\varphi,\mathbf W}$ of Proposition~\ref{prop:transport_certificate} with $\mathbf W=r^2\mathbf P^{-1}$ and $\mathbf Y=-\gamma\mathbf W$. Its distance, its minimizing geodesic from the reference, and its recovery field follow in closed form,
\begin{equation*}
\begin{gathered}
    d_{\matM}^2(\mathbf{x},0)=\varphi(\mathbf{x})^\top\mathbf P\varphi(\mathbf{x}),
    \qquad
    \bfc^\star(s;\mathbf{x})=\varphi^{-1}\bigl(s\,\varphi(\mathbf{x})\bigr),
    \\
    F_{\rec}(\mathbf{x})=-\gamma\,\mathbf J(\mathbf{x})^{-1}\varphi(\mathbf{x}),
\end{gathered}
\end{equation*}
and the recovery distance decays at exactly the rate $\gamma$.

For the deployed mode, take a constant aggressive drift in the latent coordinates, $\dot{\mathbf{z}}=\mathbf{v}_{\new}$, whose pullback $F_{\new}=\mathbf J^{-1}\mathbf v_{\new}$ is generated in~\eqref{eq:planar_controlled_plant} by $u_{\new}=\mathbf v_{\new}-8\mathbf z$. Left alone, this drift carries the state out of the safe set, as an aggressive behavior would, and its prediction $\mathbf{x}_{\new}(t)=\varphi^{-1}(\varphi(\mathbf{x}_0)+t\mathbf{v}_{\new})$ is analytic. In the plotted example, $\beta=1.10$, $\gamma=1/4$, $r=1.05$, $\mathbf{x}_0=(-0.35,0.25)$, and $\mathbf{v}_{\new}=(0.90,0.25)$.

The backup-CBF comparison uses the seed's recovery input $u_b(\mathbf z)=-(8+\gamma)\mathbf z$, whose flow is $e^{-\gamma s}\mathbf z$, as its backup policy. Its rollout set collects the states from which this policy reaches $\T_0$ within the horizon $T$ while respecting the input bound, and the standard backup-CBF rollout construction~\cite{chen2021backup} reduces to
\begin{equation*}
    \mathcal B_{T}:=\varphi^{-1}\!\left(
    \frac{8}{8+\gamma}K
    \cap\{\mathbf z:\mathbf{z}^\top\mathbf P\mathbf{z}
    \le r^2 e^{2\gamma T}\}\right).
\end{equation*}
Here $\frac{8}{8+\gamma}K$ enforces backup-input admissibility and the ellipsoid enforces terminal membership. The seed tube $\T_0$ is invariant under this policy, which uses at most $98.8\%$ of the input bound on it.

Every certified tube lies in $\mathcal V_{\rm HJ}$, since its states remain in $\mathcal S$ under admissible recovery inputs, so synthesis enforces safety against an inscribed polygon of the kernel. Let $\mathcal P_K$ be the convex hull of the intersections of $\partial K$ with the 160 rays at angles $2\pi j/160$, where $j=0,\ldots,159$. Convexity gives $\mathcal P_K\subset K$. Since each edge of $\mathcal P_K$ is affine and $\varphi^{-1}$ is polynomial, every boundary segment of $\varphi^{-1}(\mathcal P_K)$ is a polynomial curve, as Remark~\ref{rem:polynomial_boundary_oracle} requires. Algorithm~\ref{alg:transport_synthesis} starts from $\Phi^0=\varphi$ with the fixed-map optimum, the maximum-area centered ellipsoid in $\mathcal P_K$, scaled by $8/(8+\gamma)$ to enforce recovery-input admissibility. Every candidate, including this initialization, is scaled by $0.9149$ before its acceptance check. The eight accepted updates give $\Phi^8=\mathcal L_{\vartheta_8}\circ\cdots\circ\mathcal L_{\vartheta_1}\circ\varphi$ and the synthesized tube $\T_{\rm syn}=\T_{\Phi^8,\mathbf W^8}$. These shears bend the plant coordinates $\mathbf z$ further, so membership in $\T_{\rm syn}$ is tested in $\Phi^8(\mathbf x)$ rather than in $\mathbf z$.

For each accepted map, the recovery mode is the exactly realized transport field $F_{\rec}=[D\Phi]^{-1}(-\gamma\Phi)$, generated in~\eqref{eq:planar_controlled_plant} by $u_{\rec}=\mathbf JF_{\rec}-8\varphi$. Proposition~\ref{prop:transport_certificate} then gives $\T_{\rm syn}$ the closed-form squared distance $q_{\Phi,\mathbf W}$ and exact decay at rate $\gamma$ after any switch, and the CIG is algebraic,
\begin{equation*}
\G_{\matM}(\mathbf{x})
=
\bigl[D\Phi(\mathbf{x})^\top\mathbf W^{-1}\Phi(\mathbf{x})\bigr]^\top
\bigl(F_{\rec}(\mathbf{x})-F_{\new}(\mathbf{x})\bigr).
\end{equation*}

Remark~\ref{rem:polynomial_boundary_oracle} and numerical interval verification certify all 160 edges at all nine retained iterates, the initialization and the eight accepted shears. Their areas are nondecreasing, their boundary lower bounds are at least $1.0039$, and their recovery inputs use at most $99.0\%$ of the input bound, so $\T_{\rm syn}\subseteq\varphi^{-1}(\mathcal P_K)\subseteq\mathcal V_{\rm HJ}$. For switching, with $\lambda=0.35$ and $\mathbf R=0$, the deployed input uses at most $80.0\%$ of its bound throughout the first certified interval, and numerical interval sign checks show that the switching function is negative before $\tau^\star=0.7015$ and positive afterward through the first exit at $\tau_{\rm exit}=1.7097$. Hence $\tau^\star$ is the unique global minimizer of~\eqref{eq:constrained_switching_problem}, and Proposition~\ref{prop:latest_certified_switch} and Theorem~\ref{thm:root_finding_certified_recoverability} then give pathwise state and input feasibility and exponential recovery.

The synthesized tube has area $7.579$, improving by $107.3\%$ over the seed and $4.7\%$ over the input-admissible fixed-map optimum while attaining $82.9\%$ of the exact HJ viable area. Figure~\ref{fig:planar_car_recoverability}(a) shows this growth from $\T_0$, with $\T_{\rm syn}$ also exceeding the largest centered $\mathbf P$-ellipsoid $\mathcal E_c$. Figure~\ref{fig:planar_car_recoverability}(b) compares $\T_{\rm syn}$ with the exact kernel and the backup-CBF set. At the displayed horizon $T=1.25$, the area of $\T_{\rm syn}$ is $31.7\%$ larger than that of $\mathcal B_{T}$, and $\mathcal B_{T}$ needs the longer horizon $T=2.292$ to match it. Figure~\ref{fig:planar_car_recoverability}(c) follows every set under its recovery flow as each contracts to $\mathbf{x}_r=0$. Unlike the backup-CBF set, $\T_{\rm syn}$ is horizon-independent and supplies an explicit convergence rate. Outside the special structure of this plant, HJ viability requires solving a partial differential equation on a grid and backup-CBF membership requires a recovery rollout, while membership in $\T_{\rm syn}$ remains a single algebraic test. Taken together, these results show that synthesized recoverability certificates approach the exact viability kernel while keeping the closed-form test that makes them usable online.

\section{Autonomous Racing at the Limits of Handling}
\label{sec:racecar}

We now deploy the framework on a full-scale autonomous racecar at the limits of handling in the rain, a regime where vehicle dynamics models are least reliable. The quadrotor and planar examples evaluated recoverability certificates and switching programs directly, but a full-scale vehicle does not permit this at control rates. Onboard the racecar, the architecture of Fig.~\ref{fig:architecture} must run in real time using only the predictions its autonomy stack already produces. This section relaxes the exact evaluations of Section~\ref{sec:scheduling} into algebraic approximations computed onboard, and shows experimentally that they suffice to guarantee recoverability in practice.

\subsection{Considerations for Real-Time Hardware Deployment}
\label{sec:relaxations}

This subsection derives algebraic onboard evaluations, pairing each approximation with an element of the theory that already accounts for its error. Evaluating the recovery energy \eqref{eq:monitored_recovery_energy} and its gradient requires solving a geodesic boundary value problem at every control cycle. Evaluating the finite-insertion objective \eqref{eq:root_switching_objective} and its derivative requires integrating the recovery flow and the state-transition matrix \eqref{eq:recovery_stm} for every candidate switching time. Each of these issues must be addressed to enable deployment.

To this end, we adapt our approach to autonomous racing by applying our recoverability framework in transverse error coordinates. The transverse error and its rate determine whether a racecar can hold its raceline, so they are the natural coordinates in which to certify recovery to the conservative raceline. Let $\boldsymbol\zeta(\mathbf{x},t)\in\R^2$ collect the transverse error $e_\perp$ and its rate $\dot e_\perp$ relative to the conservative raceline, so that $\boldsymbol\zeta=0$ on the recovery reference $\mathbf{x}_r(t)$, and let $\Pi(\mathbf{x},t):=\partial\boldsymbol\zeta/\partial\mathbf{x}\in\R^{2\times n}$. On the recovery contraction region $\T$ of Definition~\ref{def:recovery_contraction_metric}, the deployed and recovery modes induce closed transverse vector fields $F^\zeta_\alpha:=\partial_t\boldsymbol\zeta+\Pi F_\alpha$ for $\alpha\in\{\new,\rec\}$, meaning that each $F^\zeta_\alpha$ depends on the state only through $\boldsymbol\zeta$. We choose a constant metric $M_\zeta\in\mathbb S^{2}_{++}$ on this transverse state, so that, as in the transport certificates of Section~\ref{subsec:contractive_transport}, the minimizing geodesic is a straight segment and the transverse recovery energy and its gradient are exact and algebraic,
\begin{equation}
    \mathcal V_{\loc}
    =
    \frac12\,\boldsymbol\zeta^\top M_\zeta\,\boldsymbol\zeta,
    \qquad
    \rho_{\loc}
    =
    \Pi^\top M_\zeta\,\boldsymbol\zeta,
    \label{eq:pullback_energy}
\end{equation}
and, with the transverse mode discrepancy $\Delta F^\zeta:=F^\zeta_{\rec}-F^\zeta_{\new}=\Pi\Delta F$, the transverse CIG is
\begin{equation}
    \G_{\loc}
    :=
    \rho_{\loc}^\top\Delta F
    =
    \boldsymbol\zeta^\top M_\zeta\,\Delta F^\zeta.
    \label{eq:G_loc_expanded}
\end{equation}

On the monitored tube defined by $\mathcal V_{\loc}\le E_{\max}$, recovery is certified by a nonlinear tracking controller designed to make $M_\zeta$ satisfy \eqref{eq:contraction_metric_condition} for the induced field $F^\zeta_{\rec}$ with rate $\gamma$ with respect to the conservative raceline. The region $\T$ is then forward invariant, the physical preimage of the monitored tube lies within the track corridor $\mathcal S$, and the recovery input remains admissible there. Thus, Theorem~\ref{thm:contractive_recoverability} applies to the transverse system. For deployment, we set $M_\zeta=\operatorname{diag}(0.5,1.5)$ and $E_{\max}=0.5$. Hence the monitored tube condition $\mathcal V_{\loc}\le E_{\max}$ is the ellipse
\begin{equation*}
    0.5\zeta_1^2+1.5\zeta_2^2\le1,
\end{equation*}
which admits up to $1.41$~m of transverse error at zero rate and up to $0.82$~m/s of error rate at zero error.

The predictions these evaluations require are already produced onboard our racecar. Both modes are realized as receding-horizon planners over a prediction window $[t,t+T]$ with $T=5$~s. The deployed mode $F_{\new}$ tracks an aggressive raceline and the recovery mode $F_{\rec}$ tracks the conservative raceline, and each planner publishes its predicted trajectory, $\mathbf{x}_{\new}(\cdot)$ or $\mathbf{x}_{\rec}(\cdot)$, at $\sim$10~Hz as 200 sampled nodes $\tau_i$. The five-second horizon supplies the advance warning required at racing speeds. As such, the deployed prediction $\mathbf{x}_{\new}(\cdot)$, the recovery reference, and the vector fields entering $\Delta F$ are read directly from the planners, and no dedicated integration is performed onboard. At each node $\tau_i$, both $F_{\rec}$ and $F_{\new}$ are evaluated at the same deployed state $\mathbf{x}_{\new}(\tau_i)$ and time, and we abbreviate $f(\mathbf{x}_{\new}(\tau_i),\tau_i)$ as $f(\tau_i)$. The onboard CIG~\eqref{eq:G_loc_expanded} is therefore exactly the coordinate form of Definition~\ref{def:cig}. Reference and model errors enter the induced transverse dynamics as the residuals quantified by Corollary~\ref{cor:infeasible_reference} and Theorem~\ref{thm:robust_recoverability}.

\begin{figure*}[t]
    \centering
    \includegraphics[width=0.9\textwidth]{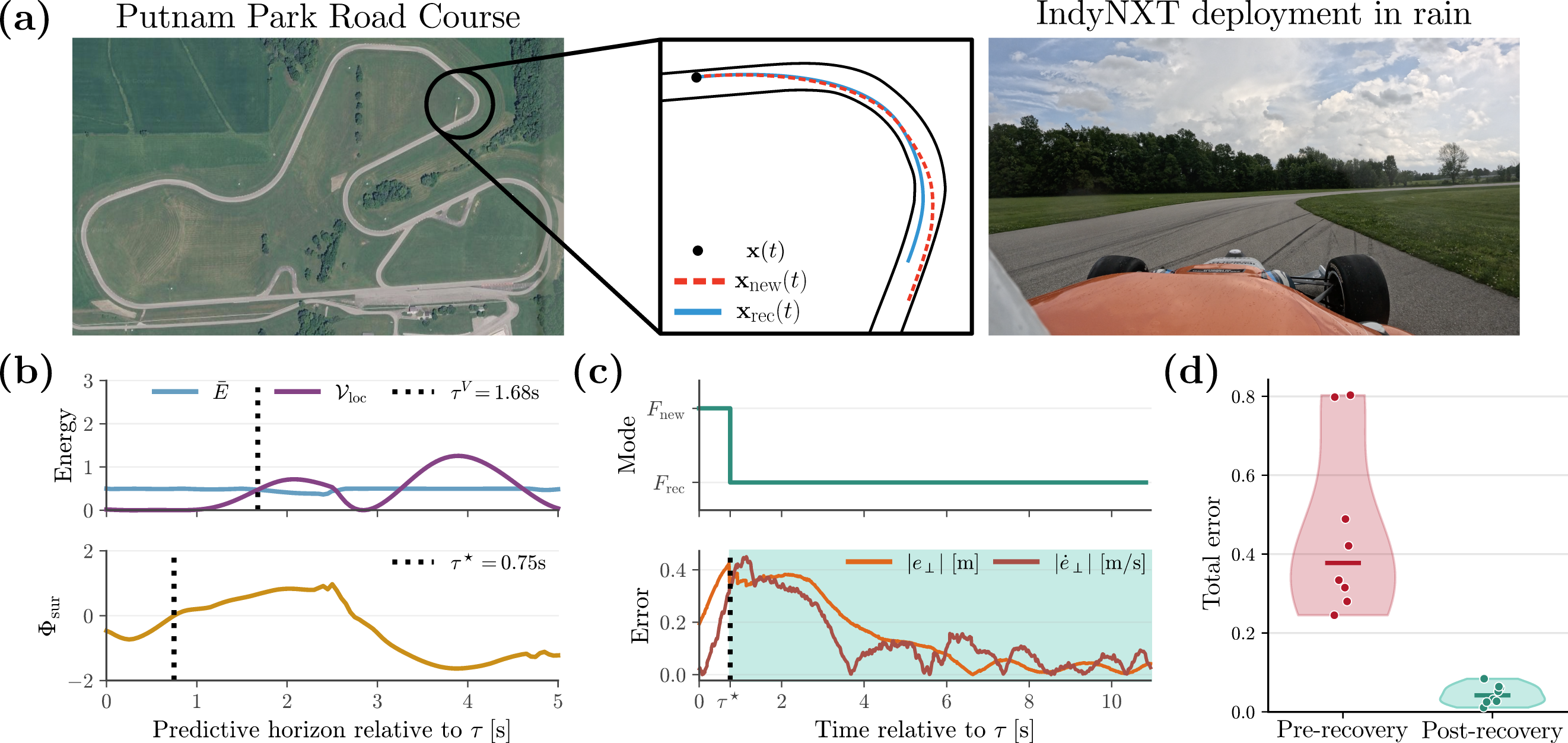}
    \caption{Certified recovery mode scheduling deployed on a full-scale autonomous racecar using the architecture of Fig.~\ref{fig:architecture}.
    (a) The vehicle is an IndyNXT racecar deployed at the Putnam Park Road Course in the rain. The inset shows the current state $\mathbf{x}(t)$ together with the aggressive prediction $\mathbf{x}_{\new}(t)$ and the recovery prediction $\mathbf{x}_{\rec}(t)$ through the upcoming corner.
    (b) A single switching decision along the predictive horizon, with times measured from the planner update. The monitored energy $\mathcal V_{\loc}$ first violates the effective threshold $\bar E$ at $\tau^V=1.68$~s, and the switching function $\Phi_{\mathrm{sur}}$ enters its tolerance band at $\tau^\star=0.75$~s, which Algorithm~\ref{alg:onboard} selects as the earliest certified candidate ahead of the violation. 
    (c) The executed recovery. The active mode steps from $F_{\new}$ to $F_{\rec}$ at $\tau^\star$, and the transverse error $|e_\perp|$ and its rate $|\dot e_\perp|$ peak near the switch and decay once the recovery mode engages, as Theorem~\ref{thm:contractive_recoverability} certifies.
    (d) Error statistics across eight certified recoveries from the deployment. Each post-recovery total error $\|\boldsymbol\zeta\|_1$ is below every pre-switch value.}
    \label{fig:car_deployment}
\end{figure*}

On hardware, a switch between modes cannot be executed instantaneously, since the recovery planner and its tracking controller take over the vehicle during a mode switch of finite duration, which we model by the insertion duration $\lambda=0.5$~s. Corollary~\ref{cor:finite_insertion_objective_consistency} already expands the switching objective to first order in this duration. Dropping the $o(\lambda)$ remainder, with $\mathcal V_{\loc}$ in place of $\mathcal V_{\matM}$, removes the recovery endpoint $\mathbf{y}_\lambda(\tau)$ from~\eqref{eq:root_switching_objective} and gives the surrogate objective
\begin{equation}
\begin{aligned}
    J_{\mathrm{sur}}(\tau)
    :=
    \mathcal V_{\loc}(\mathbf{x}_{\new}(\tau),&\tau)
    +
    \lambda\,\G_{\loc}(\mathbf{x}_{\new}(\tau),\tau)
    \\
    &+
    \frac12\|u_{\rec}(\tau)-u_{\new}(\tau)\|_{\mathbf R}^2,
\end{aligned}
\label{eq:surrogate_objective}
\end{equation}
in which $\lambda\G_{\loc}$ is the first-order effect of the mode switch and $\mathbf R=\operatorname{diag}(10^{-4},10^{-4},10^{2})$ weights the thrust, brake, and steering-angle rates. The switching function $\Phi_{\mathrm{sur}}(\tau):=\dd J_{\mathrm{sur}}(\tau)/\dd\tau$ is the slope of $J_{\mathrm{sur}}$ along the horizon, so a minimizing switch sits where it turns from negative to positive, as in Fig.~\ref{fig:car_deployment}(b). It is the surrogate counterpart of the exact switching function~\eqref{eq:switching_function_derivative}, in which the recovery-endpoint terms are replaced by $\lambda\dot\G_{\loc}$,
\begin{equation}
\begin{aligned}
    \Phi_{\mathrm{sur}}(\tau)
    =
    \dot{\mathcal V}_{\loc,\new} &(\tau)
    +
    \lambda\,\dot\G_{\loc}(\tau)
    \\
    &+
    \bigl(u_{\rec}-u_{\new}\bigr)^\top\mathbf R\bigl(\dot u_{\rec}-\dot u_{\new}\bigr),
\end{aligned}
\label{eq:phi_sur_closed_form}
\end{equation}
where $\dot{\mathcal V}_{\loc,\new}=\boldsymbol\zeta^\top M_\zeta F^\zeta_{\new}$ is the transverse instance of the energy rate~\eqref{eq:energy_rate}, and $\dot\G_{\loc}$ is the transverse instance of~\eqref{eq:Gdot},
\begin{equation}
\begin{aligned}
    \dot\G_{\loc}
    ={}&
    (F^\zeta_{\new})^\top M_\zeta\,\Delta F^\zeta
    \\
    &+
    \boldsymbol\zeta^\top M_\zeta
    \bigl(\partial_t\Delta F^\zeta+(\mathbf A^\zeta_{\rec}-\mathbf A^\zeta_{\new})F^\zeta_{\new}\bigr),
\end{aligned}
\label{eq:Gdot_loc}
\end{equation}
with transverse Jacobians $\mathbf A^\zeta_\alpha:=\partial F^\zeta_\alpha/\partial\boldsymbol\zeta$. Every term of~\eqref{eq:phi_sur_closed_form} is algebraic in the transverse dynamics and quantities the planners already publish, so neither the recovery flow nor the state-transition matrix~\eqref{eq:recovery_stm} is integrated onboard.

The same mode switch determines when the monitor must act, since what matters is the energy reached once the mode switch completes. For a forecast time $\tau$, let
\begin{equation*}
    \mathcal V_\lambda(\tau)
    :=
    \mathcal V_{\loc}\bigl(\phi_{\rec}(\tau,\tau-\lambda,\mathbf{x}_{\new}(\tau-\lambda)),\tau\bigr)
\end{equation*}
be the energy at $\tau$ after a mode switch over $[\tau-\lambda,\tau]$. Applying~\eqref{eq:finite_insertion_first_order} at the switching time $\tau-\lambda$, whose continued endpoint is $\mathbf{x}_{\new}(\tau)$, and using continuity of $\G_{\loc}$ along the prediction gives $\mathcal V_\lambda(\tau)=\mathcal V_{\loc}(\mathbf{x}_{\new}(\tau),\tau)+\lambda\G_{\loc}(\mathbf{x}_{\new}(\tau),\tau)+o(\lambda)$. Requiring the state to land inside the tube after the mode switch, to first order, $\mathcal V_\lambda\le E_{\max}$, gives the onboard trigger
\begin{equation}
\begin{aligned}
    \mathcal V_{\loc}(\mathbf{x}_{\new}(\tau),\tau)
    &\le
    \bar E(\tau),
    \\
    \bar E(\tau)
    &:=
    E_{\max}
    -
    \lambda\,\G_{\loc}(\mathbf{x}_{\new}(\tau),\tau),
\end{aligned}
    \label{eq:effective_threshold}
\end{equation}
where we refer to $\bar E$ as the effective threshold. Its first forecast violation $\tau^V$ is the first forecast time at which a mode switch of duration $\lambda$ would no longer land the state inside the tube. Equivalently, the trigger compares the surrogate objective~\eqref{eq:surrogate_objective} without its input penalty with the tube budget, so the trigger and the scheduler score the same quantity. The threshold tightens where the CIG is positive and an insertion is momentarily counterproductive, and it relaxes where the CIG is negative and recovery pulls the transverse error toward the reference. Where $\G_{\loc}<0$ the threshold exceeds $E_{\max}$, so on its own it could let the forecast leave the tube without starting a search. Algorithm~\ref{alg:onboard} therefore bounds the switching-time search by $\tau_{\rec}$, the earlier of $\tau^V$ and the sampled first exit $\tau_{\rm exit}$ of~\eqref{eq:first_certified_exit}, the first node outside the tube, and selects only nodes before $\tau_{\rec}$. Every forecast exit thus starts a search, and every node up to the selected one satisfies $\mathcal V_{\loc}\le E_{\max}$, the feasibility condition of Theorem~\ref{thm:root_finding_certified_recoverability}.

Finally, the switching program is solved on the sampled horizon and recomputed at every planner update. On this grid, the stationarity condition becomes a tolerance band $|\Phi_{\mathrm{sur}}|\le\varepsilon$ with $\varepsilon=0.05$. A forecast violation is acted on only if it recurs at a consistent location across successive planner updates, while the horizon signals are filtered to reject transient predictions. Once this persistence condition is met, Algorithm~\ref{alg:onboard} selects the earliest node in the band ahead of $\tau_{\rec}$, or the lowest-cost node when the band is empty, since by Proposition~\ref{prop:switching_kkt} the minimizer of~\eqref{eq:constrained_switching_problem} may sit at an endpoint or an active constraint where the switching function need not vanish.

\begin{algorithm}[t]
\caption{Certified Recovery Mode Scheduling} 
\label{alg:onboard}
\renewcommand{\algorithmiccomment}[1]{\hfill{\footnotesize$\triangleright$~#1}}
\algsetup{indent=0.9em}
\begin{algorithmic}[1]
\REQUIRE Metric $M_\zeta$, budget $E_{\max}$, insertion duration $\lambda$, tolerance $\varepsilon$, horizon length $T$
\FOR{each planner horizon over $[t,t+T]$}
    \STATE Read the deployed and recovery predictions
    \FOR{each node $\tau_i$ sampled from the horizon}
        \STATE Evaluate $\boldsymbol\zeta(\tau_i)$, $\mathcal V_{\loc}(\tau_i)$, $\rho_{\loc}(\tau_i)$, and $\Delta F^\zeta(\tau_i)$
        \STATE Evaluate $\G_{\loc}(\tau_i)$, $J_{\mathrm{sur}}(\tau_i)$, $\Phi_{\mathrm{sur}}(\tau_i)$, and $\bar E(\tau_i)$
    \ENDFOR
    \STATE $\tau^V = \min\,\{\tau_i : \mathcal V_{\loc}(\tau_i)>\bar E(\tau_i)\}$ \COMMENT{trigger}
    \STATE $\tau_{\rm exit} = \min\,\{\tau_i : \mathcal V_{\loc}(\tau_i)>E_{\max}\}$ \COMMENT{first exit}
    \STATE $\tau_{\rec} = \min\,\{\tau^V,\tau_{\rm exit}\}$
    \IF{$\tau_{\rec}$ exists}
        \STATE $\tau^\star = \min\,\{\tau_i<\tau_{\rec} : |\Phi_{\mathrm{sur}}(\tau_i)|\le\varepsilon\}$
        \IF{$\tau^\star$ does not exist}
            \STATE $\tau^\star\in\argmin\,\{J_{\mathrm{sur}}(\tau_i) : \tau_i<\tau_{\rec}\}$
        \ENDIF
        \STATE Switch to $F_{\rec}$ at $\tau^\star$
    \ENDIF
\ENDFOR
\end{algorithmic}
\end{algorithm}

Each approximation has an explicit cost. Reference and model errors consume part of the robustness margin $\gamma r$ and enlarge the ultimate bound of Corollary~\ref{cor:infeasible_reference} to $(\bar r+\bar d)/\gamma$. The first-order insertion model incurs an $o(\lambda)$ objective error. Sampling and filtering alter only the trigger and the selected feasible candidate. Under the hypotheses of Theorem~\ref{thm:root_finding_certified_recoverability}, these costs fall only on switching optimality and the available robustness margin, while recoverability rests on the theorem's feasibility conditions alone.

\subsection{Experimental Validation at Putnam Park Road Course}
\label{sec:experiments}

We deployed the transverse recoverability certificate and scheduler on an IndyNXT autonomous racecar at the Putnam Park Road Course, racing in the rain at speeds up to 20~m/s (45~mph), with every certificate evaluation and switching decision computed onboard in real time. Rain thinned the grip that the aggressive raceline was tuned to exploit, so the deployed mode repeatedly approached the boundary of the monitored tube through corners, and the 30-minute run produced eight certified recoveries. Figure~\ref{fig:car_deployment}(a) shows the course, the conditions, and a moment from one such corner, where the deployed prediction runs wide while the recovery prediction returns to the conservative raceline.

Figures~\ref{fig:car_deployment}(b) and~\ref{fig:car_deployment}(c) walk through one of these recoveries. Ahead of a corner, the monitored energy climbs toward the effective threshold and first crosses it at $\tau^V=1.68$~s into the predictive horizon. The switching function enters its tolerance band at $\tau^\star=0.75$~s, and the monitor schedules the feasible mode switch $0.93$~s ahead of the forecast violation. The sampled switching function is negative before $\tau^\star$ and positive thereafter through $\tau^V$, so $\tau^\star$ is a local minimizer of the sampled surrogate objective. The mode steps from $F_{\new}$ to $F_{\rec}$, and the transverse error and its rate peak near the mode switch below $0.5$~m and $0.5$~m/s before decaying. Across all eight events in Fig.~\ref{fig:car_deployment}(d), the total transverse error $\|\boldsymbol\zeta\|_1$ spans $0.25$--$0.80$ immediately before switching. All post-recovery values are below $0.09$, and the pre-switch and post-recovery medians are $0.38$ and $0.04$. Between switches the deployed mode ran completely unfiltered, so the price of safety was paid only during the recoveries themselves.

The deployed raceline in this experiment carried no certificate, and the car's safety came entirely from a certified recovery mode and a monitor that scheduled each mode switch ahead of time from predictions the autonomy stack already produced. Neither ingredient is specific to racing, since many autonomy stacks already pair planners that predict their own motion with a conservative fallback behavior that can be certified offline. Recoverability lets such robots deploy learned, aggressive, or experimental behaviors with formal guarantees, by certifying a recovery strategy rather than the behavior itself.

\section{Conclusion}
\label{sec:discussion}

Throughout this work, we have highlighted the ways in which invariance-based safety certification is at odds with robots operating at the limits of their capabilities, demanding model fidelity in the regimes where models are least reliable and penalizing operation where performance lives. We introduced a contraction-theoretic framework that overcomes these limitations by certifying a robot's ability to recover rather than its inability to fail, which we term recoverability. Our framework relocates the invariance requirement from the deployed behavior to a trusted recovery mode. A robot can then run its deployed behavior unmodified and unfiltered, with no contraction, invariance, or modeling requirement, while monitoring its safety through a single closed-form evaluation. It decides ahead of time when to switch through an optimization whose feasible solutions are all safe, and its guarantee degrades gracefully under disturbances and model error. The certified envelope itself grows monotonically toward the exact viability kernel through sequential convex synthesis. In quadrotor flight past uncertain obstacles, recoverability stayed safe where a standard CBF collided and outpaced a backup-CBF filter. Onboard a full-scale autonomous racecar, it scheduled every recovery in real time at the limits of handling in the rain. Through these validations, our work demonstrates that certified safety and uncompromised performance can coexist on real hardware in conditions that defeat model-based invariance.

However, safety certification still trails what robots are capable of, and much remains to be done to close the gap. For one, our guarantee is only as strong as the recovery mode's contraction certificate, and every recovery logged in operation is a labeled sample of the recoverability boundary, so recovery certificates learned from hardware data can improve with every deployment. Additionally, the switch we schedule is one way, and a certified re-entry condition would let robots return to aggressive operation once recovery contracts the error, turning isolated recoveries into a sustained cycle of performance and retreat. Libraries of recovery behaviors with synthesized tubes, such as braking, hovering, or returning to a safe pose, would then let a robot choose where to land rather than only when to leave~\cite{dorothy2016switched}. Beyond racing and agile flight, recoverability makes safe learning on hardware possible, since a learned policy can explore and improve in the deployed slot during operation while the recovery mode keeps every episode certified. Overall, our results recast safety certification as an exit strategy rather than a constraint on behavior, an inversion we view as crucial to the long-term coexistence of formal guarantees and robots operating at their limits.

\balance
\bibliographystyle{IEEEtran}
\bibliography{refs}

\end{document}